%% file: iclr2025_conference.tex
\documentclass{article} 
\usepackage{iclr2025_conference,times}
\usepackage{graphicx}

\input{math_commands.tex}

\usepackage{booktabs}
\usepackage{hyperref}
\usepackage{url}
\usepackage{multirow}
\usepackage{graphicx}
\usepackage{caption}

\newcommand{\name}{MeshAnything}

\title{\name: Artist-Created Mesh Generation with Autoregressive Transformers}

\author{
    \small{
        Yiwen Chen$^{1,2}$\thanks{Work done during a research internship at Shanghai AI Lab.}, ~~ ~~
        Tong He$^{2\dagger}$,~~~~
        Di Huang$^{2}$,~~~~
        Weicai Ye$^{2}$,~~~~
        Sijin Chen$^{3}$, ~~~~
        Jiaxiang Tang$^{4}$
    } \\
    \textbf{
        \small{
            Xin Chen$^{5}$,~~~~~
            Zhongang Cai$^{6}$,~~~~
            Lei Yang$^{6}$,~~~~
            Gang Yu$^{7}$,~~~~
            Guosheng Lin$^{1\dagger}$,~~~~
            Chi Zhang$^{8}$\thanks{Corresponding Authors.}
        }
    }
        \\
    {\quad\quad\quad\quad\quad\normalsize $^{1}$S-Lab, Nanyang Technological University} \quad
    {\normalsize $^{2}$Shanghai AI Lab} \\
    {\quad\normalsize $^{3}$Fudan University}~~~~
    {\normalsize $^{4}$Peking University}~~
    {\normalsize $^{5}$University of Chinese Academy of Sciences}
    \\
    {\quad\quad\quad\quad\quad\quad\normalsize $^{6}$SenseTime Research}~~~~
    {\normalsize $^{7}$Stepfun}~~~~
    {\normalsize $^{8}$Westlake University}
    \\
        \tt \small \textbf{
        \quad\quad\quad\quad\quad\quad\quad\href{https://buaacyw.github.io/mesh-anything/}{https://buaacyw.github.io/mesh-anything/}
    }
    \\
}

\iclrfinalcopy 
\begin{document}

\maketitle

\input{0_abstract}

\input{1_intro}

\input{2_related}

\input{3_AMG}

\input{4_MeshArtist}
\input{5_exp}

\input{6_conclu}

\bibliography{iclr2025_conference}
\bibliographystyle{iclr2025_conference}
\clearpage
\appendix
\input{7_append}

\end{document}

%% file: math_commands.tex
\usepackage{amsmath,amsfonts,bm}

\def\eqref#1{equation~\ref{#1}}

\def\1{\bm{1}}

\DeclareMathAlphabet{\mathsfit}{\encodingdefault}{\sfdefault}{m}{sl}
\SetMathAlphabet{\mathsfit}{bold}{\encodingdefault}{\sfdefault}{bx}{n}



%% file: 0_abstract.tex
\input{figs/F1}

\begin{abstract}
Recently, 3D assets created via reconstruction and generation have matched the quality of manually crafted assets, highlighting their potential for replacement.
However, this potential is largely unrealized because these assets always need to be converted to meshes for 3D industry applications, and the meshes produced by current mesh extraction methods are significantly inferior to Artist-Created Meshes (AMs), i.e., meshes created by human artists. 
Specifically, current mesh extraction methods rely on dense faces and ignore geometric features, leading to inefficiencies, complicated post-processing, and lower representation quality.
To address these issues, we introduce \name, a model that treats mesh extraction as a generation problem, producing AMs aligned with specified shapes.
By converting 3D assets in any 3D representation into AMs, \name~can be integrated with various 3D asset production methods, thereby enhancing their application across the 3D industry.
The architecture of \name~comprises a VQ-VAE and a shape-conditioned decoder-only transformer. We first learn a mesh vocabulary using the VQ-VAE, then train the shape-conditioned decoder-only transformer on this vocabulary for shape-conditioned autoregressive mesh generation. Our extensive experiments show that our method generates AMs with hundreds of times fewer faces, significantly improving storage, rendering, and simulation efficiencies, while achieving precision comparable to previous methods.

\end{abstract}

%% file: figs/F1.tex
    \vspace{-3mm}

\begin{figure}[h]
  \centering
\includegraphics[width=\linewidth]{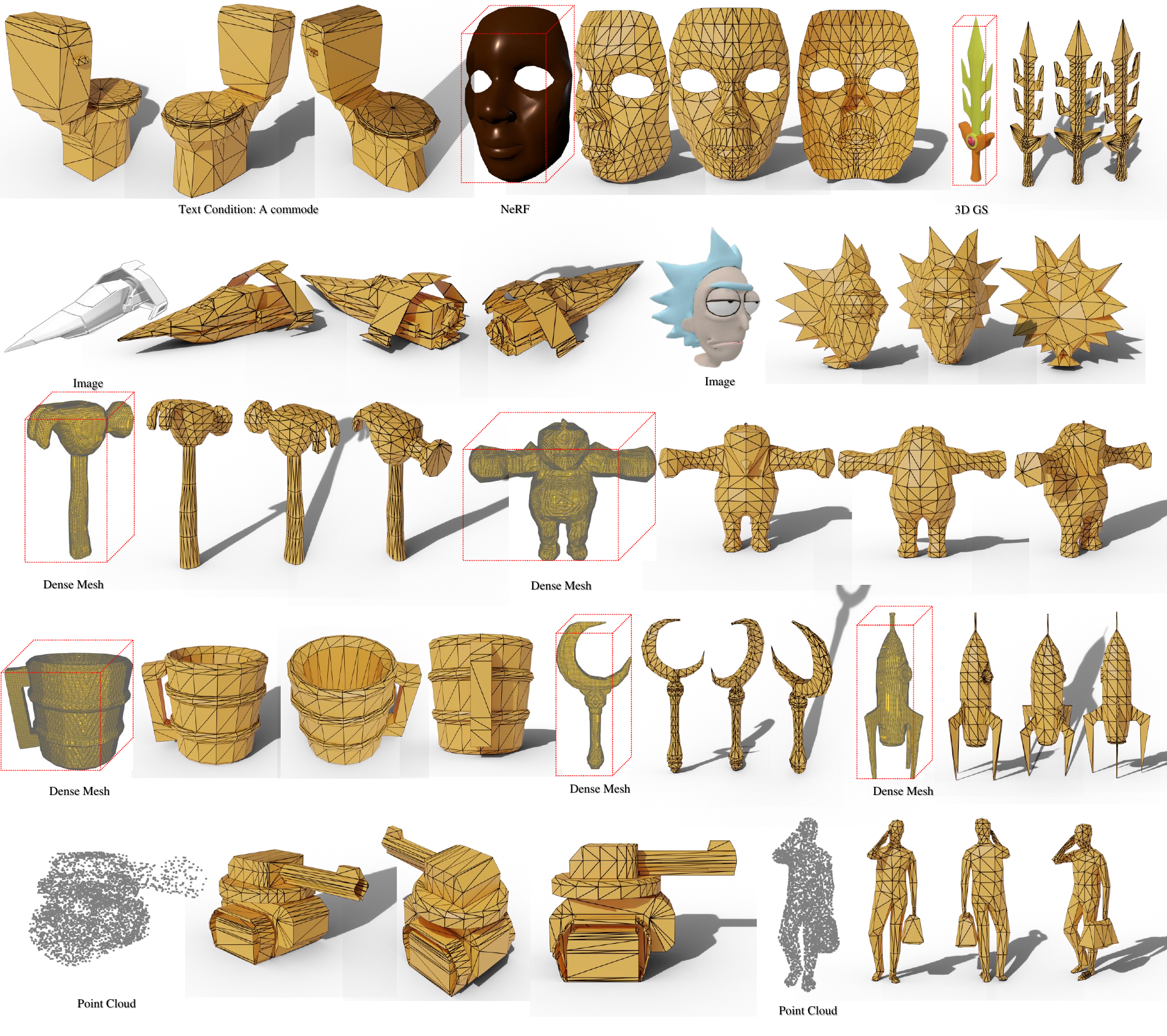}
   \caption{
 \textbf{MeshAnything converts any 3D representation into Artist-Created Meshes (AMs), i.e., meshes created by human artists.} It can be combined with various 3D asset production pipelines, such as 3D reconstruction and generation, to transform their results into AMs that can be seamlessly applied in the 3D industry.} 
   \label{fig:teaser}
\end{figure}

%% file: 1_intro.tex
\section{Introduction}

In recent years, the 3D community has experienced rapid advancements, with a variety of methods developed for automatically producing high-quality 3D assets. These methods, including 3D reconstruction~\cite{nerf,yu2021pixelnerf,barron2021mip,barron2022mipnerf360,kerbl3Dgaussians,huang20242d}, 3D generation~\cite{dreamfusion,liu2023_zero1to3,prolificdreamer,long2023wonder3d,sun2023dreamcraft3d,hong2023lrm,tang2024lgm,xu2024instantmesh,wei2024meshlrm}, and scanning~\cite{daneshmand20183d,haleem20193d,haleem2022exploring}, can produce 3D assets with shape and color quality comparable to manually created ones. The success of these methods reveals the potential to replace manually created 3D models with automatically produced ones in the 3D industry, including applications in games, movies, and the metaverse, significantly reducing time and labor costs.

However, this potential remains largely unrealized because the current 3D industry predominantly relies on mesh-based pipelines for their superior efficiency and controllability, while methods for producing 3D assets typically use alternative 3D representations to achieve optimal results across various scenarios.
Therefore, substantial efforts~\cite{lorensen1987marching,chernyaev1995marching,lorensen1998marching,shen2021deep,chen2022neural,shen2023flexible} are devoted to converting other 3D representations into meshes and have achieved some success. Meshes produced by these methods approximate the shape quality of those created by human artists, which we refer to as Artist-Created Meshes (AMs), but they still fall short in addressing the aforementioned issues. 

This is because all meshes produced by these methods~\cite{lorensen1987marching,chernyaev1995marching,lorensen1998marching,shen2021deep,chen2022neural,shen2023flexible} exhibit significantly poorer topology quality compared to AMs. As shown in Fig.~\ref{fig:remesh}, these methods rely on dense faces to reconstruct 3D shapes, completely ignoring geometric characteristics. Using these meshes in the 3D industry leads to three significant problems:
First, converted meshes typically contain several orders of magnitude more faces compared to AMs, leading to significant inefficiencies in storage, rendering, and simulation.
Moreover, the converted meshes complicate post-processing and downstream tasks in the 3D pipeline. They significantly increase the challenge for human artists in optimizing these meshes due to their chaotic and inefficient topologies.
Finally, previous methods struggle to represent sharp edges and flat surfaces, resulting in oversmoothing and bumpy artifacts as shown in Fig.~\ref{fig:remesh}.

\input{figs/F2}

In this work, we aim to solve the aforementioned issues to facilitate the application of automatically generated 3D assets in the 3D industry. As mentioned earlier, all previous methods~\cite{lorensen1987marching,chernyaev1995marching,lorensen1998marching,shen2021deep,chen2022neural,shen2023flexible} extract 3D meshes with excessively dense faces in a reconstruction manner, which inherently cannot solve these issues. Therefore, we diverge from previous approaches by formulating mesh extraction as a generation problem for the first time: we teach models to generate Artist-Created Meshes (AMs) that are aligned with the given 3D assets. The meshes generated by our methods mimic the shape and topology quality of those created by human artists. Consequently, our setting, namely Shape-Conditioned AM Generation, is fundamentally free from all previous issues, enabling seamless integration of the generated results into the 3D industry pipeline.

However, training such a model presents significant challenges. The first challenge is constructing the dataset, as we need paired shape conditions and Artist-Created Meshes (AMs) for model training. The shape condition must be efficiently derived from as many diverse 3D representations as possible to serve as a condition during inference. Additionally, it must have sufficient precision to accurately represent 3D shapes and be efficiently processed into features that can be injected into the model. After weighing the trade-offs, we chose point clouds due to their explicit and continuous representation, ease of derivation from most 3D representations, and the availability of mature point cloud encoders~\cite{qi2017pointnet,qi2017pointnet++,zhao2024michelangelo}. 

We filter out high-quality AMs from Objaverse~\cite{deitke2023objaverse,deitke2023objaversexl} and ShapeNet~\cite{chang2015shapenet}. When obtaining paired shape conditions, a naive approach would be to sample point clouds directly from AMs. However, this leads to poor results during inference because the sampled point clouds have excessive precision, while automatically produced 3D assets cannot provide point clouds of similar quality, causing a domain gap between training and inference. To address this issue, we intentionally corrupt the shape quality of AMs. We first extract the signed distance function from AMs~\cite{wang2022dual}, convert it into a coarser mesh using~\cite{lorensen1987marching}, and then sample point clouds from this coarse mesh to narrow the domain gap in shape conditions between inference and training.

Following~\cite{siddiqui2023meshgpt}, we use a VQ-VAE~\cite{van2017neural} to learn a mesh vocabulary and train a decoder-only transformer~\cite{vaswani2017attention} on this vocabulary for mesh generation. To inject shape condition, we draw inspiration from the recent success of multimodal large language models (MLLM)~\cite{wu2023multimodal,liu2024visual}, where image features encoded by pre-trained image encoders are projected into the token space of the large language models for efficient multimodal understanding.
Similarly, we treat the mesh tokens obtained from the trained VQ-VAE as the language token in LLMs and use a pre-trained encoder~\cite{zhao2024michelangelo} to encode the point clouds into shape features, which is later projected into the mesh token space. These shape tokens are placed at the beginning of the mesh token sequences, effectively serving as the shape conditions for next-token predictions. After predictions, these predicted mesh tokens are decoded back to meshes with the VQ-VAE decoder~\cite{siddiqui2023meshgpt}.

To further enhance the quality of mesh generation, we develop a novel noise-resistant decoder for robust mesh decoding. Our observation is that as the decoder in the VQ-VAE~\cite{van2017neural} is only trained with ground truth token sequences from the encoder, it could potentially lead to a domain gap when decoding the generated token sequences. To mitigate this problem, we inject the shape condition into the VQ-VAE decoder as auxiliary information for robust decoding and fine-tune it after the VQ-VAE training. This fine-tuning process involves adding noise to the mesh token sequences to simulate possible poor-quality token sequences from the decoder-only transformer, thus making the decoder robust to such poor-quality sequences.

Finally, we introduce our model, \name, trained based on the aforementioned techniques. As shown in Fig.~\ref{fig:teaser}, \name~can convert 3D assets across various 3D representations into AMs, thereby significantly facilitating their application. Furthermore, our extensive experiments demonstrate that our method generates AMs with significantly fewer faces and more refined topology, while achieving precision metrics that are close to or comparable with previous methods.

In summary, our contributions are as follows:
\begin{itemize}
    \item We highlight one important reason why current automatically produced 3D assets cannot replace those created by human artists: current methods cannot convert these 3D assets into Artist-Created Meshes (AMs). To solve this issue, we propose a novel solution called Shape-Conditioned AM Generation, which aims to generate AMs aligned with given shapes.

    \item We introduce \name~for Shape-Conditioned AM Generation. \name~can be integrated with various 3D asset production methods, converting their results into AMs to facilitate their application in the 3D industry.
    \item We develop a novel noise-resistant decoder to enhance mesh generation quality. We inject the shape condition into the decoder as auxiliary information for robust decoding and fine-tune it using noised token sequences to narrow the domain gap between training and inference.
    \item Extensive experiments demonstrate that Shape-Conditioned Mesh Generation is a more suitable setting for mesh generation, and MeshAnything significantly surpasses previous mesh generation methods.

\end{itemize}

%% file: figs/F2.tex
\begin{figure}[h]
  \centering
   \includegraphics[width=\linewidth]{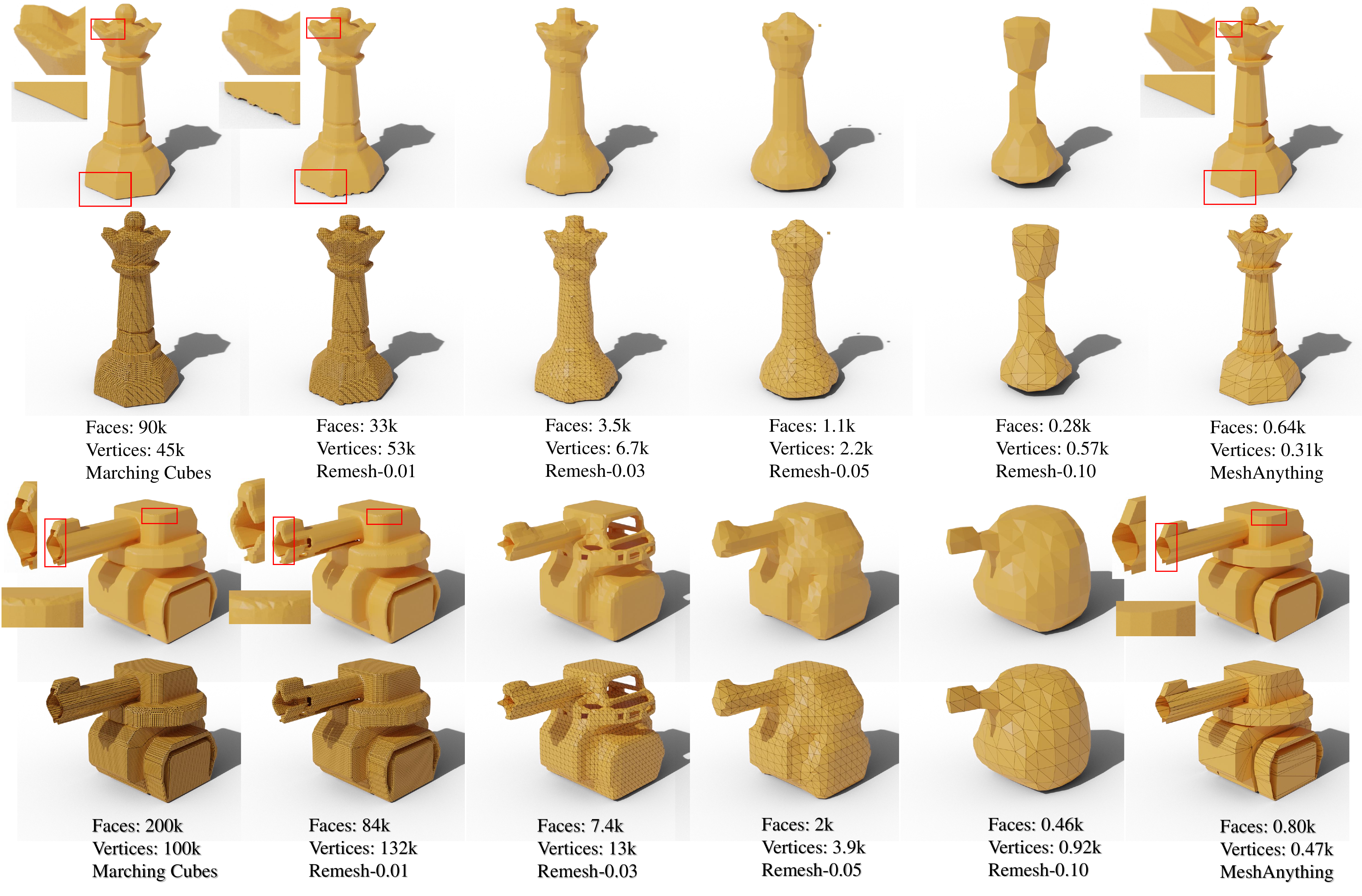}
   \caption{
\textbf{Comparison with Marching Cubes~\cite{lorensen1987marching} and Remesh~\cite{blenderremesh}.} We apply Marching Cubes and MeshAnything to ground truth shapes and then apply remeshing to the Marching Cubes results with different voxel sizes. Existing methods extract meshes in a reconstruction manner, ignoring the geometric features of the object and producing dense meshes with poor topology. These methods fundamentally fail to capture sharp edges and flat surfaces, as shown in the zoomed-in figure.}
   \label{fig:remesh}
\end{figure}

%% file: 2_related.tex
\section{Related Works}
\subsection{Mesh Extraction}
Methods for extracting meshes from 3D models are numerous and have been a subject of research for decades. Following~\cite{shen2023flexible}, we categorize these methods into two main types: Isosurface Extraction~\cite{lorensen1987marching,bloomenthal1988polygonization,chernyaev1995marching,bloomenthal1997introduction,lorensen1998marching,chen2022neural} and Gradient-Based Mesh Optimization~\cite{chen2019learning,gao2020learning,hanocka2020point2mesh,kato2018neural,dmtet,liao2018deep,shen2023flexible}. 

Traditional isosurface extraction methods~\cite{lorensen1987marching,lorensen1998marching,chernyaev1995marching,doi1991efficient,ju2002dual,schaefer2007manifold,chen2021neural,chen2022neural} focus on extracting a polygonal mesh that represents the level set of a scalar function, an area that has seen extensive study in various fields. The most popular method among them is Marching Cubes~\cite{lorensen1987marching}. It divides the space into cells, within which polygons are created to approximate the surface. Marching Cubes has been widely used for mesh extraction its robustness and simplicity. Recently,~\cite{chen2021neural} and~\cite{chen2022neural} introduce data-driven methods to determine the position of the extracted mesh based on the input field.

Transitioning to more recent developments, the advent of machine learning has ushered in new techniques for generating 3D meshes~\cite{chen2019learning,gao2020learning,hanocka2020point2mesh,kato2018neural,dmtet,liao2018deep,shen2023flexible}. This line of work explores using neural networks to generate 3D meshes, where the network parameters are optimized through gradient-based methods under specific loss functions.~\cite{dmtet} employs a differentiable Marching Tetrahedra layer for mesh extraction.
Similar to~\cite{dmtet}, ~\cite{shen2023flexible} iteratively optimizes a 3D surface mesh by representing it as the isosurface of a scalar field.

However, these approaches fundamentally differ from ours. They ignore the characteristics of the shape and inherently cannot produce meshes with efficient topology. In contrast, MeshAnything formulates mesh extraction as a generation problem for the first time, aiming to mimic human artists in mesh extraction and thereby generating Artist-Created Meshes (AMs) with hundreds of times fewer faces.

\subsection{3D Mesh Generations}
3D mesh generation can be mainly divided into two categories: generating dense meshes similar to those produced by previous mesh extraction methods, and generating Artist-Created Meshes (AMs).

The former category is currently the mainstream research focus. Methods such as~\cite{gao2022get3d,wei2024meshlrm,xu2024instantmesh} directly generate meshes in a feed-forward manner, but because they produce dense meshes with low-quality topology similar to previous mesh extraction methods, they still encounter the same issues when applied in the 3D industry.

Notably, numerous 3D generation methods~\cite{dreamfusion,tang2023make,prolificdreamer,chen2024it3d,tang2023dreamgaussian,yang2023learn,hong2023lrm,fang2023gaussianeditor,chen2023gaussianeditor,liu2024one,shi2023mvdream,li2023instant3d,chen2023text,chen2024v3d,tang2024lgm,wang2024crm,tochilkin2024triposr} can also produce meshes. These methods first generate 3D assets and then convert them to dense meshes using mesh extraction methods like~\cite{lorensen1987marching}. Consequently, they face challenges when applied to the 3D industry due to their inefficient topology.

Recently, several works have focused on the second category: generating Artist-Created Meshes(AMs)~\cite{nash2020polygen,alliegro2023polydiff,siddiqui2023meshgpt,chen2024meshxl}. Although our approach also focuses on AM generation, it fundamentally differs from these methods. Since they lack shape conditioning, these methods must simultaneously learn the complex 3D shape distribution—which typically alone requires extensive training~\cite{hong2023lrm,tang2024lgm}—and the topology distribution of AMs, leading to very challenging training processes. In contrast, our methods eliminate the challenge of learning the shape distribution, allowing the model to focus on learning the topology distribution. This not only significantly reduces training costs but also enhances the model's application value.

Among these methods, the most relevant to ours is MeshGPT~\cite{siddiqui2023meshgpt}, as we follow its architecture.~\cite{siddiqui2023meshgpt} introduced a combination of a VQ-VAE~\cite{van2017neural} and an autoregressive transformer architecture. It first learns a mesh vocabulary with the VQ-VAE and then trains the transformer on the learned vocabulary for mesh generation. However, MeshGPT's results are limited to several categories in ShapeNet. MeshGPT requires a training GPU hours similar to ours, but our method can generalize to unlimited categories in Objaverse. As shown in Fig.~\ref{ppl}, this is largely due to the difference in target complexity caused by MeshGPT needing to additionally learn the complex 3D shape distribution.

%% file: 3_AMG.tex
\input{figs/F_PPL}
\section{Shape-Conditioned AM Generation}
\label{shape-cond-mesh-gen}
In this section, we first introduce the formal formulation for Shape-Conditioned AM Generation and compare it with previous mesh generation settings~\cite{nash2020polygen,siddiqui2023meshgpt,alliegro2023polydiff}. We show that it can achieve better performance and a broader range of applications compared to the settings in previous mesh generation methods, with significantly less training effort.

Shape-Conditioned AM Generation targets to estimate a conditional distribution \( p(\mathcal{M} | \mathcal{S}) \). In this formula, \(\mathcal{M}\) refers to the Artist-Created Mesh (AM), i.e., the mesh manually modeled by human artists. \(\mathcal{S}\) refers to the 3D shape information that indicates the 3D shape to which \(\mathcal{M}\) should align. The input form of \(\mathcal{S}\) can be diverse, such as voxels or point clouds. 
Therefore, this versatility allows our method to be integrated with any 3D pipeline that outputs \(\mathcal{S}\), such as 3D reconstruction~\cite{nerf,kerbl3Dgaussians}, generation~\cite{dreamfusion,hong2023lrm}, and scanning, making these methods more efficient for the 3D industry.

Compared to existing AM generation work, they directly estimate the distribution \( p(\mathcal{M} | \mathcal{C}) \), where \(\mathcal{C}\) denotes conditions such as images, text or empty sets for unconditional generation. However, estimating \( p(\mathcal{M} | \mathcal{C}) \) requires an understanding of both the underlying shape, i.e., \(\mathcal{S}\), and complex topological structures \(\mathcal{M}\). Given this, we made the following approximation:

\begin{equation}
\label{eq:appro1}
    p(\mathcal{M} | \mathcal{C}) \approx p(\mathcal{M}, \mathcal{S} | \mathcal{C}).
\end{equation}
According to the chain rule, we have:
\begin{equation}
\label{eq:SMG1}
    p(\mathcal{M}, \mathcal{S} | \mathcal{C}) = p(\mathcal{M} | \mathcal{S}, \mathcal{C}) \cdot p(\mathcal{S} | \mathcal{C}). \\
\end{equation}
For distribution \( p(\mathcal{M} | \mathcal{S}, \mathcal{C}) \), given that \(\mathcal{S}\) is a much stronger and more direct condition than \(\mathcal{C}\), we can make the following approximation:
\begin{equation}
\label{eq:appro2}
    p(\mathcal{M} | \mathcal{S}, \mathcal{C}) \approx p(\mathcal{M} | \mathcal{S}).
\end{equation}
Combining \ref{eq:appro1}, \ref{eq:SMG1} and \ref{eq:appro2}:
\begin{equation}
\label{eq:SMG2}
    p(\mathcal{M} | \mathcal{C}) \approx p(\mathcal{M} | \mathcal{S}) \cdot p(\mathcal{S} | \mathcal{C}),
\end{equation}
in which \( p(\mathcal{M} | \mathcal{S}) \) is  the focus of our shape-conditioned mesh generation. As shown in Fig.~\ref{ppl}, estimating \( p(\mathcal{M} | \mathcal{S}) \) is much more simpler than \( p(\mathcal{M} | \mathcal{C}) \), proving that our setting is much easier to train than settings in privous methods. 
 
 As for \( p(\mathcal{S} | \mathcal{C}) \), In the 3D community, numerous large models~\cite{Rodin,tang2024lgm,xu2024instantmesh,siddiqui2023meshgpt} aim to estimate using various 3D representations and demonstrate excellent results. Besides, some single scene 3D asset production methods~\cite{nerf,kerbl3Dgaussians,barron2021mip,barron2022mipnerf360,dreamfusion,liu2023zero,sun2023dreamcraft3d} can also provide samples from this distribution.
 By integrating our framework with these existing methods, we can leverage their capabilities to enhance our mesh generation process. This integration allows for a more resource-efficient way to estimate  \( p(\mathcal{M} | \mathcal{C}) \), significantly reducing the complexity and resources required compared to previous methods.

\input{figs/pip_fig}

%% file: figs/F_PPL.tex
\begin{figure}[htbp]
    \centering
    \begin{minipage}{0.45\textwidth}
        \centering
        \includegraphics[width=\textwidth]{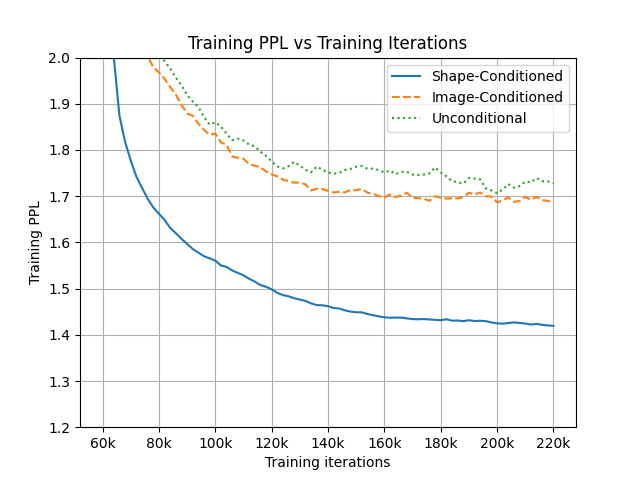}
        \caption*{(a) Training Perplexity (PPL)}
    \end{minipage}
    \hfill
    \begin{minipage}{0.45\textwidth}
        \centering
        \includegraphics[width=\textwidth]{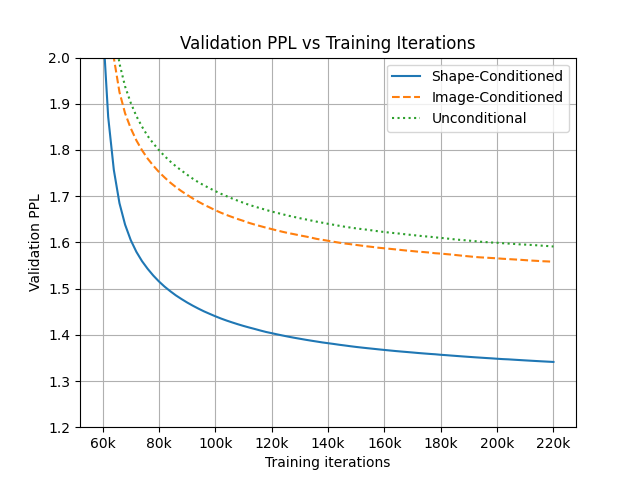}
        \caption*{(b) Validation Perplexity (PPL)}
    \end{minipage}
    \caption{\textbf{Training and validation perplexity (PPL) for the mesh generation model under different input conditions.} All models are trained with the same settings as detailed in Section~\ref{exp:imple}. The training and validation PPL of shape-conditioned mesh generation is significantly lower than that of unconditional and image-conditioned mesh generation. This indicates that the training burden of shape-conditioned mesh generation is much lower since it avoids learning the complex 3D shape distribution.}
    \label{ppl}
\end{figure}

%% file: figs/pip_fig.tex
\begin{figure}[h]
  \centering
   \includegraphics[width=\linewidth]{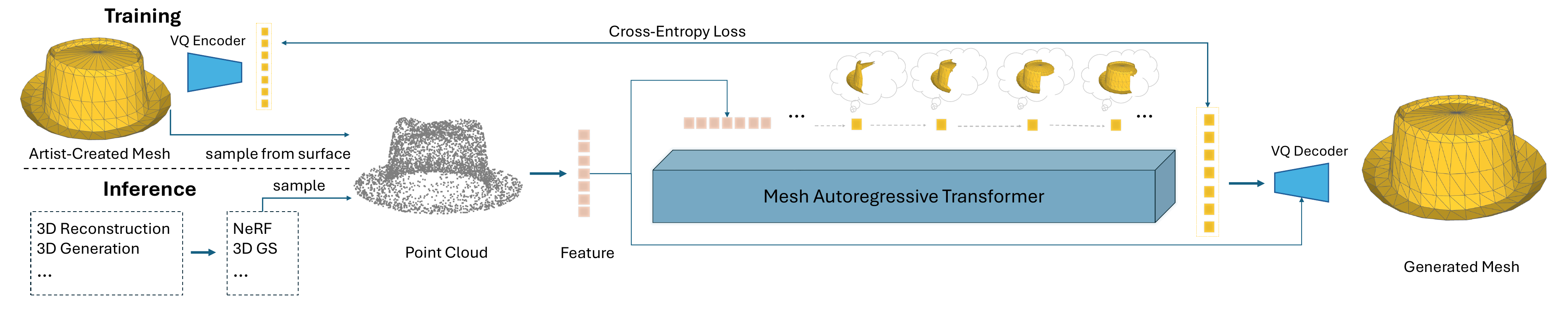}
\caption{
    \textbf{Pipeline Overview.} We introduce MeshAnything, an autoregressive transformer capable of generating Artist-Created Meshes that adhere to given 3D shapes. During training, we inject point clouds features into a decoder-only transformer and supervise it using token sequences derived from the Artist-Created meshes. After training, MeshAnything takes point clouds sampled from various 3D representations as input and generates aligned Artist-Created meshes.
}
   \label{fig:abla}
\end{figure}

%% file: 4_MeshArtist.tex
\section{Method}
In this section, we detail our shape condition strategy in Section~\ref{method:data}. After that, we provide a detailed description for~\name, which consists of a VQVAE with our newly proposed noise-resistant decoder (Section~\ref{method:vqvae}) and a shape-conditioned autoregressive transformer (Section~\ref{method:trans}).

\subsection{Shape Encoding for Conditional Generation }
\label{method:data}
We begin by describing our shape condition strategy.
\name~targets learning \( p(\mathcal{M} | \mathcal{S}) \), so we need to pair each mesh \(\mathcal{M}\) with a corresponding \(\mathcal{S}\), i.e., the shape condition. Choosing an appropriate 3D representation for \(\mathcal{S}\) is non-trivial and should satisfy the following conditions:

\begin{enumerate}
    \item It should be easily extracted from various 3D representations. This ensures that the trained models can be integrated with a wide range of 3D asset production pipelines~\cite{nerf,kerbl3Dgaussians,hong2023lrm,dreamfusion,tang2024lgm}.
    \item It should be suitable for data augmentation to prevent overfitting. To ensure the effectiveness of \(\mathcal{S}\) during training, any data augmentation applied to \(\mathcal{M}\) must be equivalently applicable to \(\mathcal{S}\).
    \item It should be efficiently and conveniently input into the model as a condition. To ensure the model comprehends the shape information and to maintain efficient training, \(\mathcal{S}\) must be easily and effectively encoded into features.
\end{enumerate}
Considering the first and second points, \(\mathcal{S}\) should be in an explicit representation. Further considering the third point, the main explicit 3D representations that can be easily encoded as features are voxels and point clouds. Both representations are suitable, but voxels typically require a high resolution to accurately represent shapes, and processing high-resolution voxels into features is computationally expensive. Additionally, voxels, being a discrete representation, are less precise for data augmentation compared to point clouds. Therefore, we chose point clouds as the representation for \(\mathcal{S}\). To enhance the expressive power of the point clouds, we also include normals into the point cloud representation.

To obtain point clouds from the ground truth mesh for training,
we could simply sample point clouds directly from the surface of \(\mathcal{M}\). However, this would create problems during inference: the surfaces of automatically generated 3D assets are often rougher than those of AMs. For example, in AMs, we would sample a series of points on a flat plane, whereas automatically generated 3D assets would have uneven surfaces, causing a domain gap between training and inference.

Therefore, we need to ensure that \(\mathcal{S}\) extracted from the ground truth \(\mathcal{M}\) during training has a similar domain to the \(\mathcal{S}\) extracted during inference.
To bring their domains closer, we intentionally construct coarse meshes from AMs.
We first extract the signed distance function from \(\mathcal{M}\) with~\cite{wang2022dual}, then convert it into a relatively coarse mesh using Marching Cubes~\cite{lorensen1987marching} to destroy the ground truth topology. Finally, we sample point cloud and its normals from the coarse mesh. This approach also helps to avoid overfitting, as AMs typically have fewer faces, and each face can often sample multiple points. The network can easily recognize the ground truth topology by determining whether the points lie on the same plane.

Since almost all 3D representations can be converted into a coarse mesh using Marching Cubes~\cite{lorensen1987marching} or sampled into point clouds, this ensures that the domain of \(\mathcal{S}\) is consistent during both training and inference. We pair the point clouds extracted as \(\mathcal{S}\) with \(\mathcal{M}\) to create a data item  \(\{(\mathcal{M}_i, \mathcal{S}_i)\}_i\) for training.

\subsection{VQ-VAE with Noise-Resistant Decoder}
\label{method:vqvae}

Following MeshGPT~\cite{siddiqui2023meshgpt}, we first train a VQ-VAE~\cite{van2017neural} to learn a vocabulary of geometric embeddings for better transformer~\cite{vaswani2017attention} learning. Different to MeshGPT, which uses graph convolutional networks~\cite{wu2019simplifying} and ResNet~\cite{he2016deep} as the encoder and decoder respectively, we employ transformers with identical structures for both the encoder and decoder. When training VQ-VAE, meshes are discretized and input as a sequence of triangle faces:
\begin{equation}
    \mathcal{M} := (f_1, f_2, f_3, \ldots, f_N), 
\end{equation}
where \(f_i\) is the coordinates of the vertices of each face, and \(N\) is the number of faces in \(\mathcal{M}\). The encoder \(E\) then extracts a feature vector for each face:
\begin{equation}
    \mathcal{Z} = (z_1, z_2, \ldots, z_N) = E(\mathcal{M}), 
\end{equation}
where \(z_i\) is the feature vector for \(f_i\).

The extracted faces are then quantized into quantized features \(\mathcal{T}\) with codebook \(\mathcal{B}\):
\begin{equation}
    \mathcal{T} = RQ(\mathcal{Z}; \mathcal{B})
\end{equation}

Finally, the reconstructed mesh is decoded from \(\mathcal{T}\) with decoder \(D\) by predicting the logits for each vertex's coordinates:
\begin{equation}
    \hat{\mathcal{M}} = D(\mathcal{Z})
\end{equation}
The VQ-VAE is trained end-to-end with cross-entropy loss on the predicted vertex coordinate logits and the commitment loss of vector quantization~\cite{van2017neural}. After the training of VQ-VAE, the encoder-decoder of VQ-VAE is treated as a tokenizer and detokenizer for autoregressive transformer training.

However, as shown in Fig.~\ref{fig:abla}, there are possible imperfections in the generation results.
To address this issue, given our setting of Shape-Conditioned AM Generation, the VQ-VAE decoder can also take the shape condition as input. Small imperfections in the token sequences generated by the transformer can potentially be corrected by a shape-aware decoder.
Therefore, after completing the vanilla VQ-VAE training, we add an additional decoder fine-tuning stage, where we inject the shape information into the transformer decoder.
Then we add random Gumbel noise to the codebook sampling logits to simulate the potential imperfections in the token sequences generated by the transformer during inference. The decoder is then updated independently with the same cross-entropy loss to train it to produce refined meshes even when facing imperfect token sequences.
Our experiments in Tab.~\ref{table:abla_noise_performance} and Tab.~\ref{table:noise_ablation} show that our method effectively enhances the decoder's noise resistance and mesh generation quality.

\subsection{Shape-Conditioned Autoregressive Transformer}
\label{method:trans}
To add shape condition to the transformer, inspired by the success of multimodal large language models~\cite{wu2023multimodal,liu2024visual,xu2023pointllm,guo2023point}, we first encode the point cloud into a fixed-length token sequence with a point cloud encoder \(\mathcal{P}\) and then concatenate it to the front of the embedding sequence from \(\mathcal{T}\) VQ-VAE as the final input embedding sequence for the transformer:
\begin{equation}
    \mathcal{T}' = \text{concat}(\mathcal{P}(\mathcal{S}), \mathcal{T})
\end{equation}
where \(\mathcal{T}'\) is the training input for the transformer.

We borrow a pretrained point encoder from~\cite{zhao2024michelangelo} and add a linear projection layer to project its output feature to the same latent space as \(\mathcal{T}\). During training, the original point encoder from~\cite{zhao2024michelangelo} is frozen; we only update the newly added projection layer and the autoregressive transformer with cross-entropy loss.

\input{tabs/user_study}

During inference, we input \(\mathcal{P}(\mathcal{S})\) to the transformer and require it to generate the subsequent sequence, \(\hat{\mathcal{T}}\). \(\hat{\mathcal{T}}\) is then input to the noise-resistant decoder to reconstruct meshes:
\begin{equation}
    \hat{\mathcal{M}} = D(\hat{\mathcal{T}})
\end{equation}
where $\hat{\mathcal{M}}$ is the final generated AM.

We use the standard next-token prediction loss to train shape-conditioned transformers. For each sequence, we add a \texttt{<bos>} token after the point cloud tokens and a \texttt{<eos>} token after the mesh tokens to identify the end of a 3D mesh.

%% file: tabs/user_study.tex
\begin{table}[h]
\caption{Comparison of Mesh Generation Methods. As shown in the left table, compared to the baseline Artist-Created Mesh Generation method, the meshes generated by MeshAnything are better aligned with human preferences. In the right table, we compare MeshAnything with mesh extraction baselines, and it received the most votes. For detailed settings, please refer to Section~\ref{exp:quantitative}.}
\centering
\begin{minipage}[b]{0.45\linewidth}
\centering
\begin{tabular}{lccc}
\toprule
\textbf{Method} & \textbf{Shape}$\uparrow$ & \textbf{Topology}$\uparrow$ \\
\midrule
PolyGen & 12.7\% & 11.1\% \\
MeshGPT & 24.1\% & 28.2\% \\
MeshAnything & \textbf{63.2\%} & \textbf{60.7\%} \\
\bottomrule
\end{tabular}
\end{minipage}
\hfill
\begin{minipage}[b]{0.45\linewidth}
\centering
\begin{tabular}{lccc}
\toprule
\textbf{Method} & \textbf{Shape}$\uparrow$ & \textbf{Topology}$\uparrow$ \\
\midrule
MarchingCubes & 38.1\% & 10.2\% \\
Shape As Points & 17.3\% & 6.2\% \\
MeshAnything & \textbf{44.6\%} & \textbf{83.6\%} \\
\bottomrule
\end{tabular}
\end{minipage}
\label{table:user}
\end{table}

%% file: 5_exp.tex
\section{Experiments}
\subsection{Data Preparation}

\textbf{Data Selection.} Existing AM generation works are limited to a few categories. However, our method targets to operate on general shapes. \name~is trained on a combined dataset of Objaverse~\cite{deitke2023objaverse} and ShapeNet~\cite{chang2015shapenet}, selected for their complementary characteristics.
We chose Objaverse because it contains a large number of AMs without category limitations. On the other hand, ShapeNet offers higher data quality within limited categories.

We filter out meshes with more than 800 faces from both datasets. Additionally, we manually filtered out low quality meshes.
Our final filtered dataset consists of 51k meshes from Objaverse and 5k meshes from ShapeNet. We randomly select 10\% of this dataset as the evaluation dataset, with the remaining 90\% used as the training set for all our experiments.

\textbf{Data Processing and Augmentation.} Following the strategies of PolyGen~\cite{nash2020polygen} and MeshGPT~\cite{siddiqui2023meshgpt}, we order faces by their lowest vertex index, then by the next lowest, and so on.  Vertices are sorted in ascending order based on their z-y-x coordinates, where z represents the vertical axis. Within each face, we permute the indices to ensure the lowest index comes first.
During training, we apply on-the-fly scaling, shifting, and rotation augmentations, normalizing each mesh to a unit bounding box from \(-0.5\) to \(0.5\).

\subsection{Implementation Details}
\label{exp:imple}
The encoder and decoder of VQ-VAE both use the encoder of BERT~\cite{devlin2018bert}, while we choose OPT-350M~\cite{zhang2022opt} as our autoregressive transformer architecture. The residual vector quantization~\cite{zeghidour2021soundstream} depth is set to 3, with a codebook size of 8,192.

Our point encoder is based on the pretrained point encoder from~\cite{zhao2024michelangelo}, which has been trained on Objaverse and thus can handle general shapes. This point encoder outputs a fixed-length token sequence of 257 tokens, with 256 tokens primarily containing shape information and an additional head token containing semantic information about the shape. We sample 4096 points for each point cloud.

The training batch size for both the VQ-VAE and the transformer is set to 8 per GPU. The VQ-VAE is trained on 8 A100 GPUs for 12 hours, after which we separately finetune the decoder part of the VQ-VAE into a noise-resistant decoder, as detailed in Section~\ref{method:vqvae}. Following this, the transformer is trained on 8 A100 GPUs for 4 days.

\input{tabs/fid}

\subsection{Qualitative Experiments}
\label{exp:qualitative}
As shown in Fig.~\ref{fig:teaser}, \name~effectively generates AMs from various 3D representations. In our experiments, we use Rodin~\cite{Rodin} as the text-to-3D and image-to-3D method, and employ~\cite{nerf} and~\cite{3dgs} as the 3D reconstruction pipeline to obtain the corresponding NeRF and Gaussian Splatting models. For additional qualitative results, please refer to ~\ref{append_exp} combined with other 3D asset production pipelines. 

\subsection{Quantitative Experiments}
From the generative model perspective, MeshAnything is a shape-conditioned mesh generation model. From the mesh extraction perspective, it extracts artist-created meshes from point clouds. Consequently, we compare MeshAnything with both types of methods. Additional experiments can be found in Appendix Section~\ref{append_exp}.

\label{exp:quantitative}
\textbf{User Study.} As shown in Tab.~\ref{table:user}, we conducted two user studies, comparing with mesh generation baselines~\cite{nash2020polygen,siddiqui2023meshgpt} and mesh extraction baselines~\cite{lorensen1987marching,peng2021shape}, respectively. The mesh generation baselines are trained on ShapeNet, and to ensure a fair comparison, we retrained them on Objaverse using the same transformer model as MeshAnything. Since the mesh generation baselines are all unconditional mesh generation methods, whereas MeshAnything is a shape-conditioned mesh generation method, we sampled shapes randomly from the evaluation set of Objaverse as inputs for MeshAnything, while for the baseline methods, we performed random sampling directly.

In the mesh extraction baseline, since our method can also be viewed as a point cloud to mesh approach, we included~\cite{peng2021shape}, a point cloud to mesh method, as a baseline. Additionally, we optimized the results from the mesh extraction baseline using the Blender remesh method~\cite{blenderremesh} to simplify the topology.

We collected $30$ results from each method and asked users to vote for the best one in terms of shape quality and topology quality. A total of $41$ users participated, providing 1,230 valid comparisons. Both user studies demonstrated the superiority of our method. The only difference between the retrained MeshGPT and MeshAnything is whether they are shape-conditioned, further proving the advantages of the shape-conditioned mesh generation setting.

\textbf{Metrics.} We follow the metric setting of~\cite{chen2022neural,siddiqui2023meshgpt}. We detail this setting in Appendix Section.~\ref{append_metrics}.

\textbf{Comparison with Mesh Generation Pipelines.} We use the same retrained models from the user study for comparison. As shown in Tab.~\ref{table:fid}, MeshAnything significantly outperforms prior methods~\cite{nash2020polygen,siddiqui2023meshgpt}, indicating that it's superior in both the shape and topology quality. Since the only difference between the retrained MeshGPT and MeshAnything is the inclusion of shape conditioning, the superior performance of MeshAnything further demonstrates that Shape-Conditioned Mesh Generation is a more suitable setting for mesh generation.

%% file: tabs/fid.tex
\begin{table}[h]
\caption{\textbf{Quantitative Comparisons with Prior Arts on Objaverse.} MeshAnything significantly outperforms prior methods across all metrics. MMD, KID are scaled by $10^3$.}

\centering
\begin{tabular}{lccccc}
\toprule
\textbf{Method} 
& \textbf{COV$\uparrow$} 
& \textbf{MMD$\downarrow$} 
& \textbf{1-NNA$\downarrow$} 
& \textbf{FID$\downarrow$} 
& \textbf{KID$\downarrow$} \\
\midrule
PolyGen & 23.2 & 6.22 & 88.2 & 48.8 & 27.7 \\
MeshGPT & 41.7 & 3.83 & 67.3 & 25.1 & 6.11 \\
MeshAnything & \textbf{53.1} & \textbf{2.72} & \textbf{55.7} & \textbf{14.5} & \textbf{1.89} \\
\bottomrule
\end{tabular}
\label{table:fid}
\end{table}

%% file: 6_conclu.tex
\section{Conclusion}

In this work, we propose a novel setting for improved mesh extraction and mesh generation, namely Shape-Conditioned Artist-Created Mesh (AM) Generation. Following this setting, we introduce \name, a model capable of generating AMs that adhere to given 3D assets. \name~can convert 3D assets in any 3D representation into AMs and thus can be integrated with diverse 3D asset production methods to facilitate their application in the 3D industry. Furthermore, we introduce a noise-resistant decoder architecture to enhance the generation quality, enabling the model to handle low-quality token sequences produced by autoregressive transformers. Lastly, extensive experiments demonstrate the superior performance of our method, highlighting its potential to scale up for 3D industry application and its advantage over previous methods.

%% file: 7_append.tex
\section{Appendix}
\input{figs/F_new}
\input{figs/F3}
\input{figs/F4}
\input{tabs/tab1}
\input{tabs/tab2}

\input{tabs/tab3}

\subsection{Metrics}
\label{append_metrics}
We follow the evaluation metric setting of~\cite{siddiqui2023meshgpt} in mesh generation experiments and the setting of~\cite{chen2022neural} in mesh extraction experiments. 

We quantitatively evaluate mesh quality by uniformly sampling 100K points from the faces of both the ground truth meshes and the predicted meshes, and then computing a set of metrics to assess various aspects of the reconstruction. 

For mesh extraction, we report the following metrics: Chamfer Distance (CD) to evaluate the overall quality of a reconstructed mesh; Edge Chamfer Distance (ECD) to assess the preservation of sharp edges by sampling points near sharp edges and corners; and Normal Consistency (NC) to evaluate the quality of the surface normals. Additionally, we report the number of mesh vertices (\#V) and the number of mesh faces (\#F). We also provide the ratio of the estimated number of vertices to the ground truth number of vertices (\#V\_R) and the same ratio for faces (\#F\_R).

For mesh generation, Coverage (COV) captures the diversity of generated meshes and is sensitive to mode dropping, but it does not reflect the quality of the results. Minimum Matching Distance (MMD) measures the average distance between the reference set and their nearest neighbors in the generated set, though it lacks sensitivity to low-quality outputs. The 1-Nearest Neighbor Accuracy (1-NNA) assesses both quality and diversity between the generated and reference sets. To evaluate topology quality, we render the ground truth meshes and generated meshes with their wireframes visualized. We then employ Frechet Inception Distance (FID) and Kernel Inception Distance (KID) on rendered images. MMD, and KID scores are scaled by a factor of $10^3$.

\subsection{Experiments}
\label{append_exp}
\textbf{Additional Qualitative Experiments}
We present more qualitative results of MeshAnything here. As shown in Fig.~\ref{fig:f_new} and Fig.~\ref{fig:qualitive}, \name~effectively generates AMs from various 3D representations. When integrated with different 3D assets production pipelines, our method effectively achieves mesh generation with diverse conditions. 

Next, Fig.~\ref{fig:qualitive} demonstrates that MeshAnything does not simply overfit but understands how to generate meshes with efficient topology that conform to the given shape. To prove this, we use manually-created meshes as ground truth and use their shapes as conditions to test whether our model can generate meshes with comparable topology. To effectively use the ground truth as conditions, we first convert them into dense meshes using Marching Cubes~\cite{lorensen1987marching} to disrupt their face structure. Then, we sample point clouds with normals from the dense meshes to serve as shape conditions. The experimental results in Fig.~\ref{fig:qualitive} show that \name~is capable of generating meshes comparable to or even surpassing those modeled by human artists, exhibiting diverse and strong 3D modeling capabilities.

\textbf{Comparison with mesh extraction baselines.} Our method is related to various mesh extraction methods~\cite{lorensen1987marching,chen2021neural,chen2022neural,shen2023flexible,peng2021shape} since we also convert other 3D representations into meshes. However, it is important to note that previous approaches are reconstruction-like methods that produce dense meshes, while our approach is generative, creating Artist-Created Meshes (AMs) that are significantly more complex to produce than dense meshes. Therefore, strictly speaking, our method cannot be considered the same as these reconstruction-based mesh extraction methods. The main purpose of this comparison is to use these mesh extraction methods as a reference for evaluating the quality of the meshes generated by MeshAnything in terms of shape. We compare MeshAnything with \cite{lorensen1987marching,shen2023flexible,peng2021shape}. Among these, MarchingCubes is the most popular mesh extraction method, FlexiCubes represents the state-of-the-art in mesh extraction, and Shape as Points is the leading method for extracting mesh from point cloud.

We also combined these methods with the remesh technique to test whether they could significantly reduce the number of faces while maintaining shape quality. We used Blender Remesh in voxel mode~\cite{blender,blenderremesh}, specifically using Blender version 4.1, as the remesh method. Since our evaluation dataset includes non-watertight meshes, we first extract the signed distance fields (SDF) of all ground truth meshes using~\cite{wang2022dual}, which can handle non-watertight meshes. We then apply Marching Cubes with a resolution of 128 on these SDFs. Next, we apply Blender remesh~\cite{blenderremesh} with different voxel sizes to the Marching Cubes results, as both the remesh method and our approach are capable of simplifying topology. Additionally, the Marching Cubes result is used as the shape condition input to \name~to obtain our results. The settings of~\cite{shen2023flexible} and~\cite{peng2021shape} follow their papers.

As shown in Tab.~\ref{table:marching_cube_remesh}, we found that these methods require hundreds of times more faces to achieve results comparable to our method. Comparing (a), (g), (i) and (k), our method lags in Chamfer Distance (CD) and Normal Consistency (NC), mainly due to our method's inherent failure cases as a generative model, which makes it less robust than these reconstruction-based mesh extraction methods. When comparing with remesh methods, we observe that they incur a high cost to achieve a face count similar to ours. Comparing (f) and (k), we find that even when remesh methods achieve a comparable face count, the number of vertices is still several times higher than ours, indicating that the topology efficiency of remesh methods is far inferior to ours, as they completely ignore the shape characteristics of the 3D assets. It's important to note that the metrics in mesh etraction can only indicate the quality of shape alignment, which do not effectively reflect the topological advantages of our method. Additionally, we surprisingly find that our method can produce results with fewer faces than the ground truth, demonstrating that \name~is not overfitting to the data but instead learns an efficient topology representation, occasionally surpassing the ground truth meshes.

\textbf{Ablations on Noise-Resistant Conditional Decoder.} We perform ablation experiments to verify the effectiveness of the Noise-Resistant Decoder. We begin with a VQ-VAE trained without any noise or conditioning. We then perform ablation between two settings: one where the decoder remains unchanged and unaware of the shape condition, and another where the shape condition is injected into the transformer, as described in Section~\ref{method:vqvae}. Next, we randomly sample a noise from gumbel distribution and add it to codebook sampling logits during the vector quantization process to simulate the potential low-quality token sequences generated by the transformer. We control the noise level by scaling the added noise.

\input{tabs/noise}

After training both models for enough epochs, we test their performance to the same level of noise.  As shown in Tab.~\ref{table:abla_noise_performance}, as the intensity of the added noise increases, the Noise-Resistant Decoder with shape condition clearly achieves better reconstruction results. This indicates that the shape condition helps the decoder identify and correct imperfections in the input token sequences. 

Next, we verify whether the Noise-Resistant Decoder indeed enhances the transformer's performance during inference. The test method used dense meshes derived from corrupted GT meshes as the condition for generating new meshes. The generated meshes were then assessed for shape alignment with the conditional shape. As shown in Tab.~\ref{table:noise_ablation}, the model with Noise-Resistant Decoder achieved better results.

\textbf{Experiments on the Impact of Input Point Cloud Quality on Generated Results.} MeshAnything takes point clouds as input, and its robustness to point cloud quality determines its versatility across various applications. We design two experiments to evaluate its tolerance to input point cloud quality: First, keeping the other evaluation settings unchanged, we apply Gaussian noise to the input point cloud coordinates and normals. Specifically, for each point, we randomly sample Gaussian noise from a standard distribution, scale it by a noise factor, and add it to the point's coordinates. The same approach is applied to the normals, but normalization is applied after adding the noise. Second, we use Rodin's generation result as the ground truth mesh, sample point clouds from this mesh as input, and evaluate the deviation between the generated result and the ground truth.

As shown in Tab.~\ref{table:point}, MeshAnything did not experience a significant performance drop in (a) and (b), demonstrating resilience to noise in the point cloud, with a noticeable performance decrease only in (c). It is important to note that the input point cloud is normalized to the range [-1,1], and the noise scale in (c) is already quite large. The experiment in (d) further demonstrates that MeshAnything can tolerate generated point clouds and effectively integrate with 3D generation models.

\subsection{Limitations}
Our method cannot generate meshes that exceed the maximum face count limit, so it cannot convert large scenes and particularly complex objects into meshes. Additionally, due to its generative nature, our method is not as stable as reconstruction-based mesh extraction methods like~\cite{lorensen1987marching,shen2023flexible}.

\subsection{Social Impact}
Our method points to a promising approach for the automatically generation of Artist-Created Meshes, which has the potential to significantly reduce labor costs in the 3D industry, thereby facilitating advancements in industries such as gaming, film, and the metaverse. However, the reduced cost of obtaining 3D Artist-Created meshes could also lead to potential criminal activities.

%% file: figs/F_new.tex
\begin{figure}[h]
  \centering
   \includegraphics[width=\linewidth]{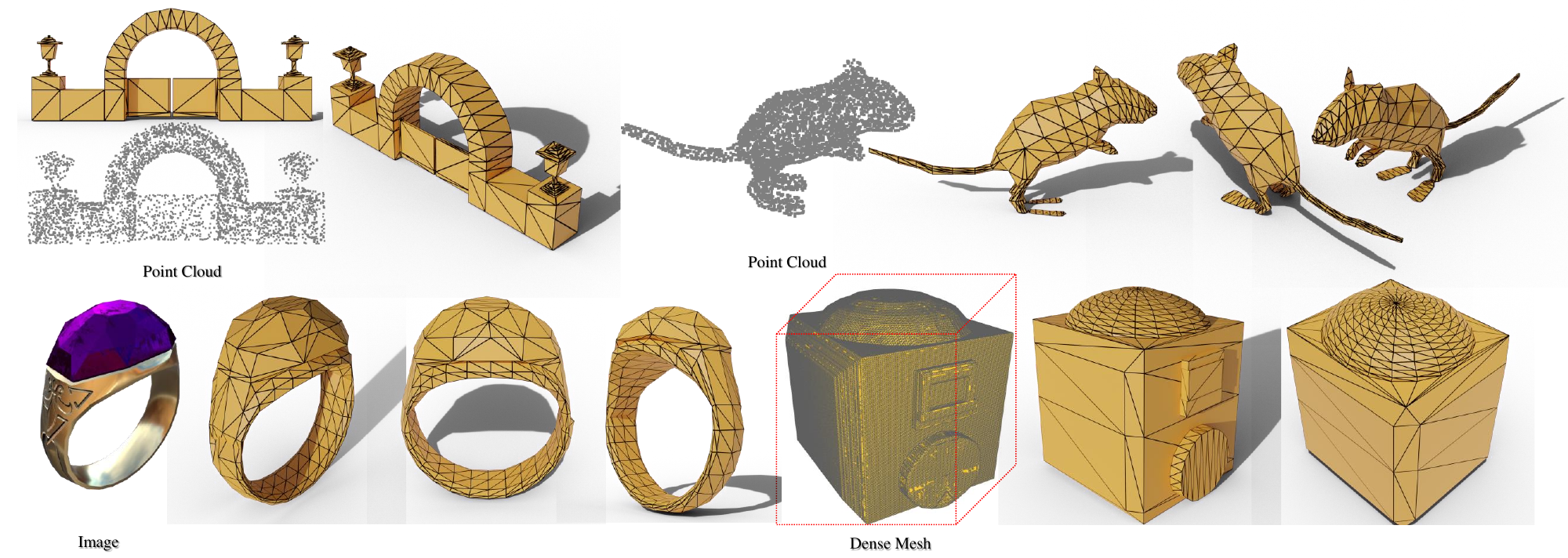}
\caption{
    \textbf{Additional qualitative results of MeshAnything.} As shown, MeshAnything can be integrated with various 3D production pipelines to achieve highly controllable mesh generation.
}
   \label{fig:f_new}
\end{figure}

%% file: figs/F3.tex
\begin{figure}[h]
  \centering
   \includegraphics[width=\linewidth]{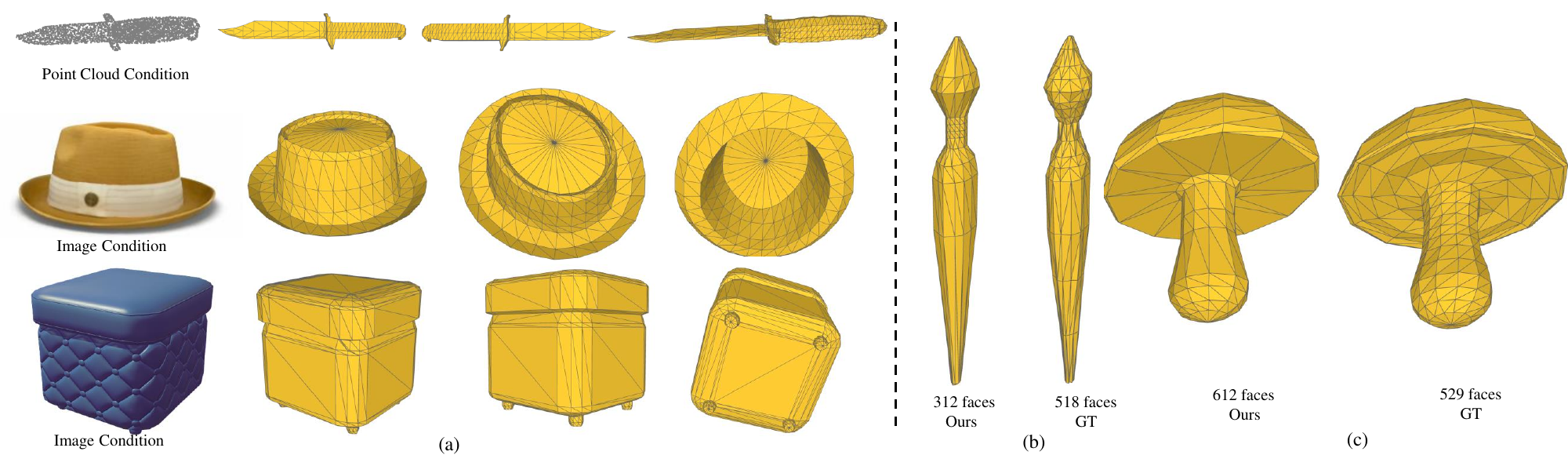}
   \caption{
\textbf{Qualitative Results.} (a) further demonstrates our capability to achieve highly controllable mesh generation when combined with 3D asset production pipelines. Besides, we compare our reseults with ground truth in (b) and (c). In (b), \name~generates meshes with better topology and fewer faces than the ground truth. In (c), we produce meshes with a completely different topology while achieving a similar shape, proving that our method does not simply overfit but understands how to construct meshes using efficient topology.
   }
   \label{fig:qualitive}

\end{figure}

%% file: figs/F4.tex
\begin{figure}[h]
  \centering
   \includegraphics[width=\linewidth]{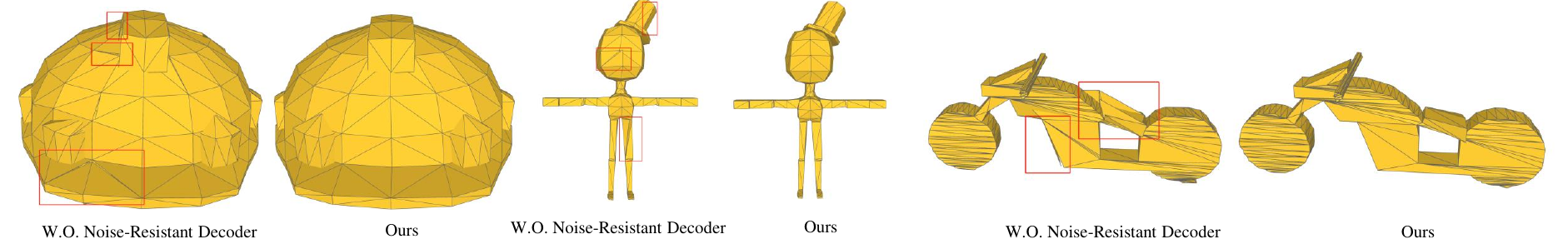}
   \caption{
 \textbf{Ablation on Noise-Resistant Decoder.} The decoder-only transformer may generate low-quality token sequences, and the decoder of VQ-VAE would typically produce flawed meshes based on these sequences. In contrast, our Noise-Resistant Decoder, aided by shape conditions, has the ability to resist these low-quality token sequences, producing higher-quality meshes.}
   \label{fig:abla}
\end{figure}

%% file: tabs/tab1.tex
\begin{table}[h!]
  \caption{Reconstruction Performance under Different Noise Levels with and without Noise-Resistant~(NR) Decoder. Please refer to~\ref{append_metrics} for metrics explanation.}
  \centering
  \begin{tabular}{ccccccccc}
    \toprule
    \multirow{2}{*}{Noise Level} 
    & \multicolumn{2}{c}{\textbf{CD}\small($\times 10^{-2}$)$\downarrow$} 
    & \multicolumn{2}{c}{\textbf{ECD}\small($\times 10^{-2}$)$\downarrow$} 
    & \multicolumn{2}{c}{\textbf{NC}$\uparrow$} \\
    \cmidrule(lr){2-3} \cmidrule(lr){4-5} \cmidrule(lr){6-7}
    & W/O NR & W/ NR 
    & W/O NR & W/ NR 
    & W/O NR & W/ NR \\
    
    \midrule
    0.0 & 0.011 & \textbf{0.007} & 0.035 & \textbf{0.023} & 0.987 & \textbf{0.993} \\
    0.1 & 0.187 & \textbf{0.028} & 0.613 & \textbf{0.138} & 0.973 & \textbf{0.991} \\
    0.5 & 1.167 & \textbf{0.639} & 2.538 & \textbf{1.329} & 0.964 & \textbf{0.981} \\
    1.0 & 2.131 & \textbf{1.798} & 4.317 & \textbf{2.316} & 0.952 & \textbf{0.969} \\
    \bottomrule
  \end{tabular}
\label{table:abla_noise_performance}
\end{table}

%% file: tabs/tab2.tex
\begin{table}[h]
\caption{Ablation on Noise-Resistant~(NR) Decoder for the Quality of Mesh Generation.}
\centering
\begin{tabular}{lcccccc}
\toprule
\textbf{Method} 
& \textbf{CD$\downarrow$} 
& \textbf{ECD$\downarrow$} 
& \textbf{NC}$\uparrow$ \\

& \small($\times 10^{-2}$)
& \small($\times 10^{-2}$)
&  \\
\midrule
W/O NR & 2.423 & 6.414 & 0.883 \\
W/ NR & \textbf{2.256} & \textbf{6.245} & \textbf{0.902} \\
\bottomrule
\end{tabular}
\label{table:noise_ablation}
\end{table}

%% file: tabs/tab3.tex
\begin{table}[h]
\caption{Quantitative evaluation with mesh extraction baselines. MC, FC, SAP refer to Marching Cubes~\cite{lorensen1987marching}, FlexiCubes~\cite{shen2023flexible}, and Shape As Points~\cite{peng2021shape}, respectively. Please refer to~\ref{append_metrics} for metrics explanation.}

\centering
\begin{tabular}{lccccccc}
\hline
\textbf{Method} & \textbf{CD$\downarrow$} & \textbf{ECD$\downarrow$} & \textbf{NC$\uparrow$} & \textbf{\#V$\downarrow$} & \textbf{\#F$\downarrow$} & \textbf{V\_R$\downarrow$} & \textbf{F\_R$\downarrow$} \\
 & \small($\times 10^{-2}$) & \small($\times 10^{-2}$) & 
 & \small($\times 10^{3}$) & \small($\times 10^{3}$) &  &  \\

\hline
(a) Marching Cubes & 1.532 & 6.733 & 0.954 & 73.22 & 146.0 & 440.2 & 462.2 \\
(b) MC+Remesh (0.005) & 2.174 & 7.813 & 0.912 & 127.8 & 167.9 & 748.1 & 534.6 \\
(c) MC+Remesh (0.010) & 2.083 & 7.578 & 0.929 & 39.01 & 41.78 & 225.4 & 132.3 \\
(d) MC+Remesh (0.030) & 2.915 & 8.329 & 0.863 & 5.848 & 4.410 & 34.38 & 14.05 \\
(e) MC+Remesh (0.050) & 4.179 & 8.138 & 0.814 & 2.299 & 1.538 & 13.64 & 4.920 \\
(f) MC+Remesh (0.100) & 7.312 & 10.771 & 0.748 & 0.625 & 0.359 & 3.735 & 1.149 \\
(g) FC & \textbf{1.190} & \textbf{6.121} & \textbf{0.967} & 59.12 & 121.1 & 378.2 & 391.1 \\
(h) FC+Remesh (0.010) & 1.861 & 6.940 & 0.933 & 37.98 & 40.19 & 205.5 & 124.2 \\
(i) SAP & 1.771 & 7.112 & 0.939 & 79.12 & 152.3 & 481.2 & 489.3 \\
(j) SAP+Remesh (0.010) & 2.367 & 7.862 & 0.925 & 39.17 & 42.87 & 239.1 & 136.6 \\
(k) \textbf{\textit{\name}} & 2.256 & 6.245 & 0.902 & \textbf{0.172} & \textbf{0.318} & \textbf{0.888} & \textbf{0.871} \\
\hline
\end{tabular}
\vspace{-3mm}
\label{table:marching_cube_remesh}
\end{table}

%% file: tabs/noise.tex
\begin{table}[h]
\caption{Experiments on the Impact of Input Point Cloud Quality on Generated Results.}

\centering
\begin{tabular}{lccccccc}
\hline
\textbf{Method} & \textbf{CD$\downarrow$} & \textbf{ECD$\downarrow$} & \textbf{NC$\uparrow$} & \textbf{\#V$\downarrow$} & \textbf{\#F$\downarrow$} & \textbf{V\_R$\downarrow$} & \textbf{F\_R$\downarrow$} \\
 & \small($\times 10^{-2}$) & \small($\times 10^{-2}$) & 
 & \small($\times 10^{3}$) & \small($\times 10^{3}$) &  &  \\

\hline
(a) Noise scale 0.005 & 2.351 & 6.412 & 0.897 & 0.175 & 0.321 & 0.895 & 0.880 \\
(b) Noise scale 0.020 & 2.980 & 6.970 & 0.881 & 0.180 & 0.330 & 0.901 & 0.910 \\
(c) Noise scale 0.050 & 4.910 & 8.556 & 0.755 & \textbf{0.162} & \textbf{0.284} & \textbf{0.811} & \textbf{0.802} \\
(d) Rodin & 2.552 & 6.622 & 0.833 & 0.185 & 0.342 & 0.919 & 0.923 \\
(e) \textbf{\textit{\name}} & \textbf{2.256} & \textbf{6.245} & \textbf{0.902} & 0.172 & 0.318 & 0.888 & 0.871 \\

\hline
\end{tabular}
\vspace{-3mm}
\label{table:point}
\end{table}

%% file: iclr2025_conference.bbl
\begin{thebibliography}{78}
\providecommand{\natexlab}[1]{#1}
\providecommand{\url}[1]{\texttt{#1}}
\expandafter\ifx\csname urlstyle\endcsname\relax
  \providecommand{\doi}[1]{doi: #1}\else
  \providecommand{\doi}{doi: \begingroup \urlstyle{rm}\Url}\fi

\bibitem[Alliegro et~al.(2023)Alliegro, Siddiqui, Tommasi, and Nie{\ss}ner]{alliegro2023polydiff}
Antonio Alliegro, Yawar Siddiqui, Tatiana Tommasi, and Matthias Nie{\ss}ner.
\newblock Polydiff: Generating 3d polygonal meshes with diffusion models.
\newblock \emph{arXiv preprint arXiv:2312.11417}, 2023.

\bibitem[Barron et~al.(2021)Barron, Mildenhall, Tancik, Hedman, Martin-Brualla, and Srinivasan]{barron2021mip}
Jonathan~T Barron, Ben Mildenhall, Matthew Tancik, Peter Hedman, Ricardo Martin-Brualla, and Pratul~P Srinivasan.
\newblock Mip-nerf: A multiscale representation for anti-aliasing neural radiance fields.
\newblock In \emph{Proceedings of the IEEE/CVF International Conference on Computer Vision}, pp.\  5855--5864, 2021.

\bibitem[Barron et~al.(2022)Barron, Mildenhall, Verbin, Srinivasan, and Hedman]{barron2022mipnerf360}
Jonathan~T. Barron, Ben Mildenhall, Dor Verbin, Pratul~P. Srinivasan, and Peter Hedman.
\newblock Mip-nerf 360: Unbounded anti-aliased neural radiance fields.
\newblock \emph{CVPR}, 2022.

\bibitem[{Blender Development Team}(2024)]{blenderremesh}
{Blender Development Team}.
\newblock Blender (version 4.1.0) [computer software], 2024.
\newblock Available from \url{https://docs.blender.org/manual/en/latest/modeling/modifiers/generate/remesh.html}.

\bibitem[Bloomenthal(1988)]{bloomenthal1988polygonization}
Jules Bloomenthal.
\newblock Polygonization of implicit surfaces.
\newblock \emph{Computer Aided Geometric Design}, 5\penalty0 (4):\penalty0 341--355, 1988.

\bibitem[Bloomenthal \& Bajaj(1997)Bloomenthal and Bajaj]{bloomenthal1997introduction}
Jules Bloomenthal and Chandrajit Bajaj.
\newblock \emph{Introduction to implicit surfaces}.
\newblock Morgan Kaufmann, 1997.

\bibitem[Chang et~al.(2015)Chang, Funkhouser, Guibas, Hanrahan, Huang, Li, Savarese, Savva, Song, Su, et~al.]{chang2015shapenet}
Angel~X Chang, Thomas Funkhouser, Leonidas Guibas, Pat Hanrahan, Qixing Huang, Zimo Li, Silvio Savarese, Manolis Savva, Shuran Song, Hao Su, et~al.
\newblock Shapenet: An information-rich 3d model repository.
\newblock \emph{arXiv preprint arXiv:1512.03012}, 2015.

\bibitem[Chen et~al.(2024{\natexlab{a}})Chen, Chen, Pang, Zeng, Cheng, Fu, Yin, Wang, Wang, Zhang, et~al.]{chen2024meshxl}
Sijin Chen, Xin Chen, Anqi Pang, Xianfang Zeng, Wei Cheng, Yijun Fu, Fukun Yin, Yanru Wang, Zhibin Wang, Chi Zhang, et~al.
\newblock Meshxl: Neural coordinate field for generative 3d foundation models.
\newblock \emph{arXiv preprint arXiv:2405.20853}, 2024{\natexlab{a}}.

\bibitem[Chen et~al.(2019)Chen, Ling, Gao, Smith, Lehtinen, Jacobson, and Fidler]{chen2019learning}
Wenzheng Chen, Huan Ling, Jun Gao, Edward Smith, Jaakko Lehtinen, Alec Jacobson, and Sanja Fidler.
\newblock Learning to predict 3d objects with an interpolation-based differentiable renderer.
\newblock \emph{Advances in neural information processing systems}, 32, 2019.

\bibitem[Chen et~al.(2023{\natexlab{a}})Chen, Chen, Zhang, Wang, Yang, Wang, Cai, Yang, Liu, and Lin]{chen2023gaussianeditor}
Yiwen Chen, Zilong Chen, Chi Zhang, Feng Wang, Xiaofeng Yang, Yikai Wang, Zhongang Cai, Lei Yang, Huaping Liu, and Guosheng Lin.
\newblock Gaussianeditor: Swift and controllable 3d editing with gaussian splatting.
\newblock \emph{arXiv preprint arXiv:2311.14521}, 2023{\natexlab{a}}.

\bibitem[Chen et~al.(2024{\natexlab{b}})Chen, Zhang, Yang, Cai, Yu, Yang, and Lin]{chen2024it3d}
Yiwen Chen, Chi Zhang, Xiaofeng Yang, Zhongang Cai, Gang Yu, Lei Yang, and Guosheng Lin.
\newblock It3d: Improved text-to-3d generation with explicit view synthesis.
\newblock In \emph{Proceedings of the AAAI Conference on Artificial Intelligence}, volume~38, pp.\  1237--1244, 2024{\natexlab{b}}.

\bibitem[Chen \& Zhang(2021)Chen and Zhang]{chen2021neural}
Zhiqin Chen and Hao Zhang.
\newblock Neural marching cubes.
\newblock \emph{ACM Transactions on Graphics (TOG)}, 40\penalty0 (6):\penalty0 1--15, 2021.

\bibitem[Chen et~al.(2022)Chen, Tagliasacchi, Funkhouser, and Zhang]{chen2022neural}
Zhiqin Chen, Andrea Tagliasacchi, Thomas Funkhouser, and Hao Zhang.
\newblock Neural dual contouring.
\newblock \emph{ACM Transactions on Graphics (TOG)}, 41\penalty0 (4):\penalty0 1--13, 2022.

\bibitem[Chen et~al.(2023{\natexlab{b}})Chen, Wang, and Liu]{chen2023text}
Zilong Chen, Feng Wang, and Huaping Liu.
\newblock Text-to-3d using gaussian splatting.
\newblock \emph{arXiv preprint arXiv:2309.16585}, 2023{\natexlab{b}}.

\bibitem[Chen et~al.(2024{\natexlab{c}})Chen, Wang, Wang, Wang, and Liu]{chen2024v3d}
Zilong Chen, Yikai Wang, Feng Wang, Zhengyi Wang, and Huaping Liu.
\newblock V3d: Video diffusion models are effective 3d generators.
\newblock \emph{arXiv preprint arXiv:2403.06738}, 2024{\natexlab{c}}.

\bibitem[Chernyaev(1995)]{chernyaev1995marching}
Evgeni Chernyaev.
\newblock Marching cubes 33: Construction of topologically correct isosurfaces.
\newblock Technical report, 1995.

\bibitem[Community(2018)]{blender}
Blender~Online Community.
\newblock \emph{Blender - a 3D modelling and rendering package}.
\newblock Blender Foundation, Stichting Blender Foundation, Amsterdam, 2018.
\newblock URL \url{http://www.blender.org}.

\bibitem[Daneshmand et~al.(2018)Daneshmand, Helmi, Avots, Noroozi, Alisinanoglu, Arslan, Gorbova, Haamer, Ozcinar, and Anbarjafari]{daneshmand20183d}
Morteza Daneshmand, Ahmed Helmi, Egils Avots, Fatemeh Noroozi, Fatih Alisinanoglu, Hasan~Sait Arslan, Jelena Gorbova, Rain~Eric Haamer, Cagri Ozcinar, and Gholamreza Anbarjafari.
\newblock 3d scanning: A comprehensive survey.
\newblock \emph{arXiv preprint arXiv:1801.08863}, 2018.

\bibitem[Deitke et~al.(2023{\natexlab{a}})Deitke, Liu, Wallingford, Ngo, Michel, Kusupati, Fan, Laforte, Voleti, Gadre, et~al.]{deitke2023objaversexl}
Matt Deitke, Ruoshi Liu, Matthew Wallingford, Huong Ngo, Oscar Michel, Aditya Kusupati, Alan Fan, Christian Laforte, Vikram Voleti, Samir~Yitzhak Gadre, et~al.
\newblock Objaverse-xl: A universe of 10m+ 3d objects.
\newblock \emph{arXiv preprint arXiv:2307.05663}, 2023{\natexlab{a}}.

\bibitem[Deitke et~al.(2023{\natexlab{b}})Deitke, Schwenk, Salvador, Weihs, Michel, VanderBilt, Schmidt, Ehsani, Kembhavi, and Farhadi]{deitke2023objaverse}
Matt Deitke, Dustin Schwenk, Jordi Salvador, Luca Weihs, Oscar Michel, Eli VanderBilt, Ludwig Schmidt, Kiana Ehsani, Aniruddha Kembhavi, and Ali Farhadi.
\newblock Objaverse: A universe of annotated 3d objects.
\newblock In \emph{CVPR}, pp.\  13142--13153, 2023{\natexlab{b}}.

\bibitem[Devlin et~al.(2018)Devlin, Chang, Lee, and Toutanova]{devlin2018bert}
Jacob Devlin, Ming-Wei Chang, Kenton Lee, and Kristina Toutanova.
\newblock Bert: Pre-training of deep bidirectional transformers for language understanding.
\newblock \emph{arXiv preprint arXiv:1810.04805}, 2018.

\bibitem[Doi \& Koide(1991)Doi and Koide]{doi1991efficient}
Akio Doi and Akio Koide.
\newblock An efficient method of triangulating equi-valued surfaces by using tetrahedral cells.
\newblock \emph{IEICE TRANSACTIONS on Information and Systems}, 74\penalty0 (1):\penalty0 214--224, 1991.

\bibitem[Fang et~al.(2023)Fang, Wang, Zhang, Xie, and Tian]{fang2023gaussianeditor}
Jiemin Fang, Junjie Wang, Xiaopeng Zhang, Lingxi Xie, and Qi~Tian.
\newblock Gaussianeditor: Editing 3d gaussians delicately with text instructions.
\newblock \emph{arXiv preprint arXiv:2311.16037}, 2023.

\bibitem[Gao et~al.(2020)Gao, Chen, Xiang, Jacobson, McGuire, and Fidler]{gao2020learning}
Jun Gao, Wenzheng Chen, Tommy Xiang, Alec Jacobson, Morgan McGuire, and Sanja Fidler.
\newblock Learning deformable tetrahedral meshes for 3d reconstruction.
\newblock \emph{Advances in neural information processing systems}, 33:\penalty0 9936--9947, 2020.

\bibitem[Gao et~al.(2022)Gao, Shen, Wang, Chen, Yin, Li, Litany, Gojcic, and Fidler]{gao2022get3d}
Jun Gao, Tianchang Shen, Zian Wang, Wenzheng Chen, Kangxue Yin, Daiqing Li, Or~Litany, Zan Gojcic, and Sanja Fidler.
\newblock Get3d: A generative model of high quality 3d textured shapes learned from images.
\newblock \emph{NeurIPS}, 35:\penalty0 31841--31854, 2022.

\bibitem[Guo et~al.(2023)Guo, Zhang, Zhu, Tang, Ma, Han, Chen, Gao, Li, Li, et~al.]{guo2023point}
Ziyu Guo, Renrui Zhang, Xiangyang Zhu, Yiwen Tang, Xianzheng Ma, Jiaming Han, Kexin Chen, Peng Gao, Xianzhi Li, Hongsheng Li, et~al.
\newblock Point-bind \& point-llm: Aligning point cloud with multi-modality for 3d understanding, generation, and instruction following.
\newblock \emph{arXiv preprint arXiv:2309.00615}, 2023.

\bibitem[Haleem \& Javaid(2019)Haleem and Javaid]{haleem20193d}
Abid Haleem and Mohd Javaid.
\newblock 3d scanning applications in medical field: a literature-based review.
\newblock \emph{Clinical Epidemiology and Global Health}, 7\penalty0 (2):\penalty0 199--210, 2019.

\bibitem[Haleem et~al.(2022)Haleem, Javaid, Singh, Rab, Suman, Kumar, and Khan]{haleem2022exploring}
Abid Haleem, Mohd Javaid, Ravi~Pratap Singh, Shanay Rab, Rajiv Suman, Lalit Kumar, and Ibrahim~Haleem Khan.
\newblock Exploring the potential of 3d scanning in industry 4.0: An overview.
\newblock \emph{International Journal of Cognitive Computing in Engineering}, 3:\penalty0 161--171, 2022.

\bibitem[Hanocka et~al.(2020)Hanocka, Metzer, Giryes, and Cohen-Or]{hanocka2020point2mesh}
Rana Hanocka, Gal Metzer, Raja Giryes, and Daniel Cohen-Or.
\newblock Point2mesh: A self-prior for deformable meshes.
\newblock \emph{arXiv preprint arXiv:2005.11084}, 2020.

\bibitem[He et~al.(2016)He, Zhang, Ren, and Sun]{he2016deep}
Kaiming He, Xiangyu Zhang, Shaoqing Ren, and Jian Sun.
\newblock Deep residual learning for image recognition.
\newblock In \emph{Proceedings of the IEEE conference on computer vision and pattern recognition}, pp.\  770--778, 2016.

\bibitem[Hong et~al.(2023)Hong, Zhang, Gu, Bi, Zhou, Liu, Liu, Sunkavalli, Bui, and Tan]{hong2023lrm}
Yicong Hong, Kai Zhang, Jiuxiang Gu, Sai Bi, Yang Zhou, Difan Liu, Feng Liu, Kalyan Sunkavalli, Trung Bui, and Hao Tan.
\newblock Lrm: Large reconstruction model for single image to 3d.
\newblock \emph{arXiv preprint arXiv:2311.04400}, 2023.

\bibitem[Huang et~al.(2024)Huang, Yu, Chen, Geiger, and Gao]{huang20242d}
Binbin Huang, Zehao Yu, Anpei Chen, Andreas Geiger, and Shenghua Gao.
\newblock 2d gaussian splatting for geometrically accurate radiance fields.
\newblock \emph{arXiv preprint arXiv:2403.17888}, 2024.

\bibitem[Ju et~al.(2002)Ju, Losasso, Schaefer, and Warren]{ju2002dual}
Tao Ju, Frank Losasso, Scott Schaefer, and Joe Warren.
\newblock Dual contouring of hermite data.
\newblock In \emph{Proceedings of the 29th annual conference on Computer graphics and interactive techniques}, pp.\  339--346, 2002.

\bibitem[Kato et~al.(2018)Kato, Ushiku, and Harada]{kato2018neural}
Hiroharu Kato, Yoshitaka Ushiku, and Tatsuya Harada.
\newblock Neural 3d mesh renderer.
\newblock In \emph{Proceedings of the IEEE conference on computer vision and pattern recognition}, pp.\  3907--3916, 2018.

\bibitem[Kerbl et~al.(2023{\natexlab{a}})Kerbl, Kopanas, Leimk{\"u}hler, and Drettakis]{3dgs}
Bernhard Kerbl, Georgios Kopanas, Thomas Leimk{\"u}hler, and George Drettakis.
\newblock 3d gaussian splatting for real-time radiance field rendering.
\newblock \emph{ACM Transactions on Graphics (ToG)}, 42\penalty0 (4):\penalty0 1--14, 2023{\natexlab{a}}.

\bibitem[Kerbl et~al.(2023{\natexlab{b}})Kerbl, Kopanas, Leimk{\"u}hler, and Drettakis]{kerbl3Dgaussians}
Bernhard Kerbl, Georgios Kopanas, Thomas Leimk{\"u}hler, and George Drettakis.
\newblock 3d gaussian splatting for real-time radiance field rendering.
\newblock \emph{ACM Transactions on Graphics}, 42\penalty0 (4), July 2023{\natexlab{b}}.
\newblock URL \url{https://repo-sam.inria.fr/fungraph/3d-gaussian-splatting/}.

\bibitem[Li et~al.(2023)Li, Tan, Zhang, Xu, Luan, Xu, Hong, Sunkavalli, Shakhnarovich, and Bi]{li2023instant3d}
Jiahao Li, Hao Tan, Kai Zhang, Zexiang Xu, Fujun Luan, Yinghao Xu, Yicong Hong, Kalyan Sunkavalli, Greg Shakhnarovich, and Sai Bi.
\newblock Instant3d: Fast text-to-3d with sparse-view generation and large reconstruction model.
\newblock \emph{arXiv preprint arXiv:2311.06214}, 2023.

\bibitem[Liao et~al.(2018)Liao, Donne, and Geiger]{liao2018deep}
Yiyi Liao, Simon Donne, and Andreas Geiger.
\newblock Deep marching cubes: Learning explicit surface representations.
\newblock In \emph{Proceedings of the IEEE Conference on Computer Vision and Pattern Recognition}, pp.\  2916--2925, 2018.

\bibitem[Liu et~al.(2024{\natexlab{a}})Liu, Li, Wu, and Lee]{liu2024visual}
Haotian Liu, Chunyuan Li, Qingyang Wu, and Yong~Jae Lee.
\newblock Visual instruction tuning.
\newblock \emph{Advances in neural information processing systems}, 36, 2024{\natexlab{a}}.

\bibitem[Liu et~al.(2024{\natexlab{b}})Liu, Xu, Jin, Chen, Varma~T, Xu, and Su]{liu2024one}
Minghua Liu, Chao Xu, Haian Jin, Linghao Chen, Mukund Varma~T, Zexiang Xu, and Hao Su.
\newblock One-2-3-45: Any single image to 3d mesh in 45 seconds without per-shape optimization.
\newblock \emph{Advances in Neural Information Processing Systems}, 36, 2024{\natexlab{b}}.

\bibitem[Liu et~al.(2023{\natexlab{a}})Liu, Wu, Hoorick, Tokmakov, Zakharov, and Vondrick]{liu2023_zero1to3}
Ruoshi Liu, Rundi Wu, Basile~Van Hoorick, Pavel Tokmakov, Sergey Zakharov, and Carl Vondrick.
\newblock {Zero-1-to-3}: Zero-shot one image to 3d object.
\newblock \emph{https://arxiv.org/abs/2303.11328}, 2023{\natexlab{a}}.

\bibitem[Liu et~al.(2023{\natexlab{b}})Liu, Wu, Van~Hoorick, Tokmakov, Zakharov, and Vondrick]{liu2023zero}
Ruoshi Liu, Rundi Wu, Basile Van~Hoorick, Pavel Tokmakov, Sergey Zakharov, and Carl Vondrick.
\newblock Zero-1-to-3: Zero-shot one image to 3d object.
\newblock \emph{arXiv preprint arXiv:2303.11328}, 2023{\natexlab{b}}.

\bibitem[Long et~al.(2023)Long, Guo, Lin, Liu, Dou, Liu, Ma, Zhang, Habermann, Theobalt, et~al.]{long2023wonder3d}
Xiaoxiao Long, Yuan-Chen Guo, Cheng Lin, Yuan Liu, Zhiyang Dou, Lingjie Liu, Yuexin Ma, Song-Hai Zhang, Marc Habermann, Christian Theobalt, et~al.
\newblock Wonder3d: Single image to 3d using cross-domain diffusion.
\newblock \emph{arXiv preprint arXiv:2310.15008}, 2023.

\bibitem[Lorensen \& Cline(1987)Lorensen and Cline]{lorensen1987marching}
William~E Lorensen and Harvey~E Cline.
\newblock Marching cubes: A high resolution 3d surface construction algorithm.
\newblock \emph{ACM siggraph computer graphics}, 21\penalty0 (4):\penalty0 163--169, 1987.

\bibitem[Lorensen \& Cline(1998)Lorensen and Cline]{lorensen1998marching}
William~E Lorensen and Harvey~E Cline.
\newblock Marching cubes: A high resolution 3d surface construction algorithm.
\newblock In \emph{Seminal graphics: pioneering efforts that shaped the field}, pp.\  347--353. 1998.

\bibitem[Mildenhall et~al.(2020)Mildenhall, Srinivasan, Tancik, Barron, Ramamoorthi, and Ng]{nerf}
Ben Mildenhall, Pratul~P. Srinivasan, Matthew Tancik, Jonathan~T. Barron, Ravi Ramamoorthi, and Ren Ng.
\newblock Nerf: Representing scenes as neural radiance fields for view synthesis.
\newblock In \emph{ECCV}, 2020.

\bibitem[Nash et~al.(2020)Nash, Ganin, Eslami, and Battaglia]{nash2020polygen}
Charlie Nash, Yaroslav Ganin, SM~Ali Eslami, and Peter Battaglia.
\newblock Polygen: An autoregressive generative model of 3d meshes.
\newblock In \emph{International conference on machine learning}, pp.\  7220--7229. PMLR, 2020.

\bibitem[Peng et~al.(2021)Peng, Jiang, Liao, Niemeyer, Pollefeys, and Geiger]{peng2021shape}
Songyou Peng, Chiyu Jiang, Yiyi Liao, Michael Niemeyer, Marc Pollefeys, and Andreas Geiger.
\newblock Shape as points: A differentiable poisson solver.
\newblock \emph{Advances in Neural Information Processing Systems}, 34:\penalty0 13032--13044, 2021.

\bibitem[Poole et~al.(2023)Poole, Jain, Barron, and Mildenhall]{dreamfusion}
Ben Poole, Ajay Jain, Jonathan~T. Barron, and Ben Mildenhall.
\newblock Dreamfusion: Text-to-3d using 2d diffusion.
\newblock In \emph{The Eleventh International Conference on Learning Representations, {ICLR} 2023, Kigali, Rwanda, May 1-5, 2023}. OpenReview.net, 2023.
\newblock URL \url{https://openreview.net/pdf?id=FjNys5c7VyY}.

\bibitem[Qi et~al.(2017{\natexlab{a}})Qi, Su, Mo, and Guibas]{qi2017pointnet}
Charles~R Qi, Hao Su, Kaichun Mo, and Leonidas~J Guibas.
\newblock Pointnet: Deep learning on point sets for 3d classification and segmentation.
\newblock In \emph{CVPR 2017}, pp.\  652--660, 2017{\natexlab{a}}.

\bibitem[Qi et~al.(2017{\natexlab{b}})Qi, Yi, Su, and Guibas]{qi2017pointnet++}
Charles~Ruizhongtai Qi, Li~Yi, Hao Su, and Leonidas~J Guibas.
\newblock Pointnet++: Deep hierarchical feature learning on point sets in a metric space.
\newblock \emph{Advances in neural information processing systems}, 30, 2017{\natexlab{b}}.

\bibitem[Schaefer et~al.(2007)Schaefer, Ju, and Warren]{schaefer2007manifold}
Scott Schaefer, Tao Ju, and Joe Warren.
\newblock Manifold dual contouring.
\newblock \emph{IEEE Transactions on Visualization and Computer Graphics}, 13\penalty0 (3):\penalty0 610--619, 2007.

\bibitem[Shen et~al.(2021{\natexlab{a}})Shen, Gao, Yin, Liu, and Fidler]{dmtet}
Tianchang Shen, Jun Gao, Kangxue Yin, Ming-Yu Liu, and Sanja Fidler.
\newblock Deep marching tetrahedra: a hybrid representation for high-resolution 3d shape synthesis.
\newblock In \emph{Advances in Neural Information Processing Systems (NeurIPS)}, 2021{\natexlab{a}}.

\bibitem[Shen et~al.(2021{\natexlab{b}})Shen, Gao, Yin, Liu, and Fidler]{shen2021deep}
Tianchang Shen, Jun Gao, Kangxue Yin, Ming-Yu Liu, and Sanja Fidler.
\newblock Deep marching tetrahedra: a hybrid representation for high-resolution 3d shape synthesis.
\newblock \emph{Advances in Neural Information Processing Systems}, 34:\penalty0 6087--6101, 2021{\natexlab{b}}.

\bibitem[Shen et~al.(2023)Shen, Munkberg, Hasselgren, Yin, Wang, Chen, Gojcic, Fidler, Sharp, and Gao]{shen2023flexible}
Tianchang Shen, Jacob Munkberg, Jon Hasselgren, Kangxue Yin, Zian Wang, Wenzheng Chen, Zan Gojcic, Sanja Fidler, Nicholas Sharp, and Jun Gao.
\newblock Flexible isosurface extraction for gradient-based mesh optimization.
\newblock \emph{ACM Transactions on Graphics (TOG)}, 42\penalty0 (4):\penalty0 1--16, 2023.

\bibitem[Shi et~al.(2023)Shi, Wang, Ye, Long, Li, and Yang]{shi2023mvdream}
Yichun Shi, Peng Wang, Jianglong Ye, Mai Long, Kejie Li, and Xiao Yang.
\newblock Mvdream: Multi-view diffusion for 3d generation.
\newblock \emph{arXiv preprint arXiv:2308.16512}, 2023.

\bibitem[Siddiqui et~al.(2023)Siddiqui, Alliegro, Artemov, Tommasi, Sirigatti, Rosov, Dai, and Nie{\ss}ner]{siddiqui2023meshgpt}
Yawar Siddiqui, Antonio Alliegro, Alexey Artemov, Tatiana Tommasi, Daniele Sirigatti, Vladislav Rosov, Angela Dai, and Matthias Nie{\ss}ner.
\newblock Meshgpt: Generating triangle meshes with decoder-only transformers.
\newblock \emph{arXiv preprint arXiv:2311.15475}, 2023.

\bibitem[Sun et~al.(2023)Sun, Zhang, Shao, Wang, Liu, Xie, and Liu]{sun2023dreamcraft3d}
Jingxiang Sun, Bo~Zhang, Ruizhi Shao, Lizhen Wang, Wen Liu, Zhenda Xie, and Yebin Liu.
\newblock Dreamcraft3d: Hierarchical 3d generation with bootstrapped diffusion prior.
\newblock \emph{arXiv preprint arXiv:2310.16818}, 2023.

\bibitem[Tang et~al.(2023{\natexlab{a}})Tang, Ren, Zhou, Liu, and Zeng]{tang2023dreamgaussian}
Jiaxiang Tang, Jiawei Ren, Hang Zhou, Ziwei Liu, and Gang Zeng.
\newblock Dreamgaussian: Generative gaussian splatting for efficient 3d content creation.
\newblock \emph{arXiv preprint arXiv:2309.16653}, 2023{\natexlab{a}}.

\bibitem[Tang et~al.(2024)Tang, Chen, Chen, Wang, Zeng, and Liu]{tang2024lgm}
Jiaxiang Tang, Zhaoxi Chen, Xiaokang Chen, Tengfei Wang, Gang Zeng, and Ziwei Liu.
\newblock Lgm: Large multi-view gaussian model for high-resolution 3d content creation.
\newblock \emph{arXiv preprint arXiv:2402.05054}, 2024.

\bibitem[Tang et~al.(2023{\natexlab{b}})Tang, Wang, Zhang, Zhang, Yi, Ma, and Chen]{tang2023make}
Junshu Tang, Tengfei Wang, Bo~Zhang, Ting Zhang, Ran Yi, Lizhuang Ma, and Dong Chen.
\newblock Make-it-3d: High-fidelity 3d creation from a single image with diffusion prior.
\newblock \emph{arXiv preprint arXiv:2303.14184}, 2023{\natexlab{b}}.

\bibitem[Team(2024)]{Rodin}
Deemos Team.
\newblock Deemos rodin.
\newblock \url{https://hyperhuman.deemos.com/rodin}, 2024.

\bibitem[Tochilkin et~al.(2024)Tochilkin, Pankratz, Liu, Huang, Letts, Li, Liang, Laforte, Jampani, and Cao]{tochilkin2024triposr}
Dmitry Tochilkin, David Pankratz, Zexiang Liu, Zixuan Huang, Adam Letts, Yangguang Li, Ding Liang, Christian Laforte, Varun Jampani, and Yan-Pei Cao.
\newblock Triposr: Fast 3d object reconstruction from a single image.
\newblock \emph{arXiv preprint arXiv:2403.02151}, 2024.

\bibitem[Van Den~Oord et~al.(2017)Van Den~Oord, Vinyals, et~al.]{van2017neural}
Aaron Van Den~Oord, Oriol Vinyals, et~al.
\newblock Neural discrete representation learning.
\newblock \emph{Advances in neural information processing systems}, 30, 2017.

\bibitem[Vaswani et~al.(2017)Vaswani, Shazeer, Parmar, Uszkoreit, Jones, Gomez, Kaiser, and Polosukhin]{vaswani2017attention}
Ashish Vaswani, Noam Shazeer, Niki Parmar, Jakob Uszkoreit, Llion Jones, Aidan~N Gomez, {\L}ukasz Kaiser, and Illia Polosukhin.
\newblock Attention is all you need.
\newblock \emph{Advances in neural information processing systems}, 30, 2017.

\bibitem[Wang et~al.(2022)Wang, Liu, and Tong]{wang2022dual}
Peng-Shuai Wang, Yang Liu, and Xin Tong.
\newblock Dual octree graph networks for learning adaptive volumetric shape representations.
\newblock \emph{ACM Transactions on Graphics (TOG)}, 41\penalty0 (4):\penalty0 1--15, 2022.

\bibitem[Wang et~al.(2023)Wang, Lu, Wang, Bao, Li, Su, and Zhu]{prolificdreamer}
Zhengyi Wang, Cheng Lu, Yikai Wang, Fan Bao, Chongxuan Li, Hang Su, and Jun Zhu.
\newblock Prolificdreamer: High-fidelity and diverse text-to-3d generation with variational score distillation.
\newblock \emph{arXiv preprint arXiv:2305.16213}, 2023.

\bibitem[Wang et~al.(2024)Wang, Wang, Chen, Xiang, Chen, Yu, Li, Su, and Zhu]{wang2024crm}
Zhengyi Wang, Yikai Wang, Yifei Chen, Chendong Xiang, Shuo Chen, Dajiang Yu, Chongxuan Li, Hang Su, and Jun Zhu.
\newblock Crm: Single image to 3d textured mesh with convolutional reconstruction model.
\newblock \emph{arXiv preprint arXiv:2403.05034}, 2024.

\bibitem[Wei et~al.(2024)Wei, Zhang, Bi, Tan, Luan, Deschaintre, Sunkavalli, Su, and Xu]{wei2024meshlrm}
Xinyue Wei, Kai Zhang, Sai Bi, Hao Tan, Fujun Luan, Valentin Deschaintre, Kalyan Sunkavalli, Hao Su, and Zexiang Xu.
\newblock Meshlrm: Large reconstruction model for high-quality mesh.
\newblock \emph{arXiv preprint arXiv:2404.12385}, 2024.

\bibitem[Wu et~al.(2019)Wu, Souza, Zhang, Fifty, Yu, and Weinberger]{wu2019simplifying}
Felix Wu, Amauri Souza, Tianyi Zhang, Christopher Fifty, Tao Yu, and Kilian Weinberger.
\newblock Simplifying graph convolutional networks.
\newblock In \emph{International conference on machine learning}, pp.\  6861--6871. PMLR, 2019.

\bibitem[Wu et~al.(2023)Wu, Gan, Chen, Wan, and Philip]{wu2023multimodal}
Jiayang Wu, Wensheng Gan, Zefeng Chen, Shicheng Wan, and S~Yu Philip.
\newblock Multimodal large language models: A survey.
\newblock In \emph{2023 IEEE International Conference on Big Data (BigData)}, pp.\  2247--2256. IEEE, 2023.

\bibitem[Xu et~al.(2024)Xu, Cheng, Gao, Wang, Gao, and Shan]{xu2024instantmesh}
Jiale Xu, Weihao Cheng, Yiming Gao, Xintao Wang, Shenghua Gao, and Ying Shan.
\newblock Instantmesh: Efficient 3d mesh generation from a single image with sparse-view large reconstruction models.
\newblock \emph{arXiv preprint arXiv:2404.07191}, 2024.

\bibitem[Xu et~al.(2023)Xu, Wang, Wang, Chen, Pang, and Lin]{xu2023pointllm}
Runsen Xu, Xiaolong Wang, Tai Wang, Yilun Chen, Jiangmiao Pang, and Dahua Lin.
\newblock Pointllm: Empowering large language models to understand point clouds.
\newblock \emph{arXiv preprint arXiv:2308.16911}, 2023.

\bibitem[Yang et~al.(2023)Yang, Chen, Chen, Zhang, Xu, Yang, Liu, and Lin]{yang2023learn}
Xiaofeng Yang, Yiwen Chen, Cheng Chen, Chi Zhang, Yi~Xu, Xulei Yang, Fayao Liu, and Guosheng Lin.
\newblock Learn to optimize denoising scores for 3d generation: A unified and improved diffusion prior on nerf and 3d gaussian splatting.
\newblock \emph{arXiv preprint arXiv:2312.04820}, 2023.

\bibitem[Yu et~al.(2021)Yu, Ye, Tancik, and Kanazawa]{yu2021pixelnerf}
Alex Yu, Vickie Ye, Matthew Tancik, and Angjoo Kanazawa.
\newblock pixelnerf: Neural radiance fields from one or few images.
\newblock In \emph{CVPR}, pp.\  4578--4587, 2021.

\bibitem[Zeghidour et~al.(2021)Zeghidour, Luebs, Omran, Skoglund, and Tagliasacchi]{zeghidour2021soundstream}
Neil Zeghidour, Alejandro Luebs, Ahmed Omran, Jan Skoglund, and Marco Tagliasacchi.
\newblock Soundstream: An end-to-end neural audio codec.
\newblock \emph{IEEE/ACM Transactions on Audio, Speech, and Language Processing}, 30:\penalty0 495--507, 2021.

\bibitem[Zhang et~al.(2022)Zhang, Roller, Goyal, Artetxe, Chen, Chen, Dewan, Diab, Li, Lin, et~al.]{zhang2022opt}
Susan Zhang, Stephen Roller, Naman Goyal, Mikel Artetxe, Moya Chen, Shuohui Chen, Christopher Dewan, Mona Diab, Xian Li, Xi~Victoria Lin, et~al.
\newblock Opt: Open pre-trained transformer language models.
\newblock \emph{arXiv preprint arXiv:2205.01068}, 2022.

\bibitem[Zhao et~al.(2024)Zhao, Liu, Chen, Zeng, Wang, Cheng, Fu, Chen, Yu, and Gao]{zhao2024michelangelo}
Zibo Zhao, Wen Liu, Xin Chen, Xianfang Zeng, Rui Wang, Pei Cheng, Bin Fu, Tao Chen, Gang Yu, and Shenghua Gao.
\newblock Michelangelo: Conditional 3d shape generation based on shape-image-text aligned latent representation.
\newblock \emph{Advances in Neural Information Processing Systems}, 36, 2024.

\end{thebibliography}
